\documentclass{article}

\usepackage{PRIMEarxiv}

\usepackage[utf8]{inputenc} 
\usepackage[T1]{fontenc}    
\usepackage{hyperref}       
\usepackage{url}            
\usepackage{booktabs}       
\usepackage{amsfonts}       
\usepackage{nicefrac}       
\usepackage{microtype}      
\usepackage{lipsum}
\usepackage{fancyhdr}       
\usepackage{graphicx}       
\graphicspath{{media/}}     

\usepackage{amsmath}
\usepackage{amssymb}
\usepackage{graphicx}
\usepackage{multirow} 
\usepackage{longtable} 
\usepackage{array} 
\usepackage{makecell}
\usepackage{threeparttable} 
\usepackage{booktabs}
\usepackage{float}
\usepackage{stfloats}
\usepackage{multirow} 
\usepackage{subfig}

\title{ElasticLaneNet: An Efficient Geometry-Flexible Approach for Lane Detection
}
\author{
  Yaxin FENG$^{1}$, Yuan LAN$^{1}$, Luchan ZHANG$^{3*}$, Yang XIANG$^{1,2}$\thanks{Corresponding author.} \\ 
$^{1}$Department of Mathematics, Hong Kong University of Science and Technology, \\ Clear Water Bay, Hong Kong SAR, China \\
$^{2}$Algorithms of Machine Learning and Autonomous Driving Research Lab, HKUST Shenzhen-Hong Kong \\ Collaborative Innovation Research Institute, Futian, Shenzhen, China \\
$^{3}$College of Mathematics and Statistics, Shenzhen University, Shenzhen 518060, China \\
\texttt{\{yfengba, ylanaa\}@connect.ust.hk}, \texttt{zhanglc@szu.edu.cn}, \texttt{maxiang@ust.hk} \\
}

\begin{document}
\maketitle

\begin{abstract}
  The task of lane detection involves identifying the boundaries of driving areas in real-time. Recognizing lanes with variable and complex geometric structures remains a challenge. In this paper, we explore a novel and flexible way of implicit lanes representation named \textit{Elastic Lane map (ELM)}, and introduce an efficient physics-informed end-to-end lane detection framework, namely, ElasticLaneNet (Elastic interaction energy-informed Lane detection Network). The approach considers predicted lanes as moving zero-contours on the flexibly shaped \textit{ELM} that are attracted to the ground truth guided by an elastic interaction energy-loss function (EIE loss). Our framework well integrates the global information and low-level features. The method performs well in complex lane scenarios, including those with large curvature, weak geometry features at intersections, complicated cross lanes, Y-shapes lanes, dense lanes, etc. We apply our approach on three datasets: SDLane, CULane, and TuSimple. The results demonstrate exceptional performance of our method, with the state-of-the-art results on the structurally diverse SDLane, achieving F1-score of 89.51, Recall rate of 87.50, and Precision of 91.61 with fast inference speed.
\end{abstract}


\section{Introduction}
\label{sec:intro}

\begin{figure}[t]
  \centering
   \includegraphics[width=0.85\linewidth]{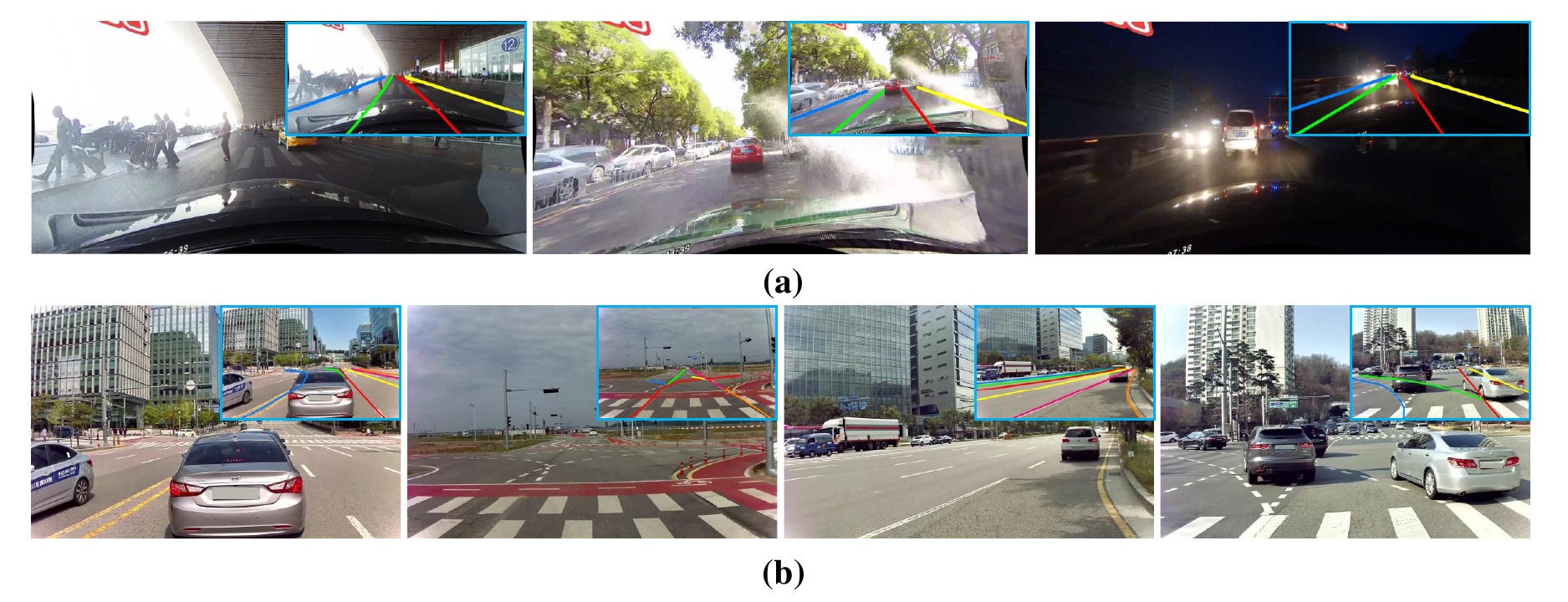}
   \caption{Illustrations of challenging cases for lane detection. (a) Weak features, including occlusion, no line or dashed lines, night/shadow/dazzle light conditions~\cite{pan2018spatial}. (b) Complex geometry lanes, such as Y-shaped (new branch, intersections, turns), dense, merging, winding driveway, etc~\cite{jin2022eigenlanes}.}
   \label{fig:intro}
\end{figure}

Artificial intelligence is widely used in autonomous driving and the Advanced Driver Assistance System (ADAS). Lane detection is a critical technique for ADAS, which helps the intelligent vehicles plan the driving maneuvers. In addition to the real-time requirement, the challenges in lane detection~\cite{qin2020ultra,jin2022eigenlanes,liu2021condlanenet,pan2018spatial} can be divided into two categories: weak lane features and complex geometry lanes; see  Fig.~\ref{fig:intro}. Weak lane features means that the lanes are not obvious, as shown in  Fig.~\ref{fig:intro} (a), which happen when the roads have no line, occlusion caused by traffic and crowd, or poor illumination with shadow, night, dazzle light, bad weather, etc. Figure~\ref{fig:intro} (b) illustrates the diversity of road geometry (complex structures and shapes) including cross roads on intersections, lanes splits and merge in the ramps, dense or large curvature lanes.

The traditional methods of lane detection~\cite{liu2010combining,zhou2010novel,hur2013multi,aly2008real} mainly use image gradient to detect the lane edge by handcraft operators, \textit{e.g.} Hough transform~\cite{liu2010combining,zhou2010novel} followed by post-processing, such as RANSAC~\cite{jiang2009new,kim2008robust}. However, these approaches are not good at coping with a large amount of complicated perceptual tasks in various real scene. Due to the rapid development of deep learning, many Learning-based lane detection methods~\cite{pan2018spatial,tabelini2021polylanenet,neven2018towards,zheng2022clrnet,liu2021condlanenet,jin2022eigenlanes} have shown ground-breaking effects on achieving end-to-end understanding on automatic driving scenes. Nonetheless, most of the methods aim at straight or low curvature lanes without forks, while the real driving scenes are rich and diverse, and the structure of lanes are various. It is necessary to explore and develop more robust lane models for these complex geometry lanes. 

\begin{figure*}[t]
  \centering
   \includegraphics[width=0.8\textwidth]{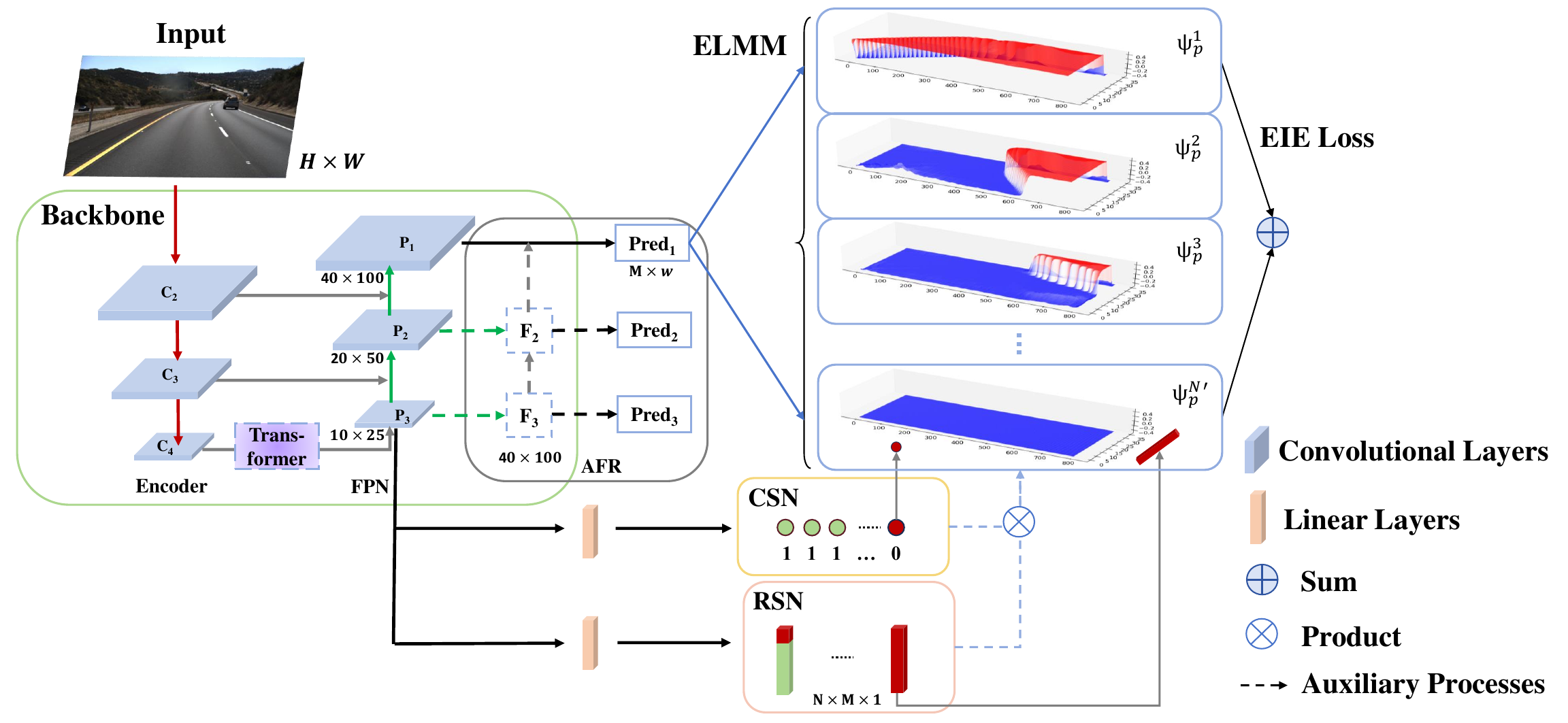}
   \caption{Overview of the proposed ElasticLaneNet architecture. The backbone is constructed as Encoder-FPN pipeline. \textit{Elastic Lane Map} Module (ELMM) is the lane detection head. Classification and range sub-networks (CSN, RSN) are jointly trained and producted to obtain the final \textit{ELM}. TB layer and AFR module are incorporated in the network. The EIE loss measures the distance between prediction and the ground truth. }
   \label{fig:networkarchitecture}
\end{figure*}

In this work, we propose a novel lane detection framework named ElasticLaneNet; see  Fig.~\ref{fig:networkarchitecture}. In our framework, lanes are modeled as open curves without width that implicitly embedded on the \textit{Elastic Lane Map (ELM)}, whose coordinates can be directly and efficiently collected row by row as zero-contours of \textit{ELM}. In our lane model, \textit{ELM} guided via elastic interaction energy loss function (EIE loss) enables better combination of global information and low-level features. These properties enable our approach overcome the drawbacks of segmentation-based lane models (regarding lanes as long thin objects with finite width, usually about 30 pixels)~\cite{pan2018spatial,hou2019learning,zheng2021resa,abualsaud2021laneaf,xu2020curvelane} and other row-wise approaches based on coarse confidence maps~\cite{liu2021condlanenet,qin2020ultra,qin2022ultra}, \textit{e.g.} class imbalance of multi-instance classification (the number of positive samples is much smaller than the number of negative samples), costly post processing and inaccurate lane points selection from multiple pixels/coarse grids on maps. In addition, implicit representations are much more geometric-flexible than parameter-based~\cite{tabelini2021polylanenet,feng2022rethinking,liu2021end} and anchor-based models~\cite{li2019line,tabelini2021keep,zheng2022clrnet} because lanes can be diversely shaped and bend with any angle less than 90$^{\circ}$ on \textit{ELM}. 


Our elastic interaction energy loss function (EIE loss) is inspired from physical properties of line defects in crystals~\cite{hirth1983theory}. It is found in~\cite{feng2023elastic} that the EIE loss can guide more accurate results of fine-scale instance in multi-scale segmentation. Here, we build a better and properer lane model motivated by this superior property of the EIE loss. As the EIE loss is an integral over the entire implicit map \textit{ELM}, our model considers each lane as a whole, and integrates the cross rows context, while segmentation mainly focuses on local features. Therefore, our model can overcome the challenge of weak lane features appearing in the current segmentation-based lane detection methods, \textit{e.g.} lane missed, disconnection or blurry~\cite{pan2018spatial,hou2019learning,zheng2021resa,abualsaud2021laneaf,xu2020curvelane}. Under the guidance of the EIE loss, the curve generated by our \textit{ELM} is smooth and connected, and well approximates to the ground truth, even when sampling points are sparse.


We evaluate our approach on three popular autonomous driving lane datasets, namely TuSimple~\cite{tusimple2017benchmark}, CULane~\cite{pan2018spatial}, and a newly proposed structure diverse data SDLane~\cite{jin2022eigenlanes}. The experimental results show that our geometric flexibility strategy can achieve SOTA on the complex dataset SDLane, as well as competitive results on the two other commonly used datasets, with fast inference speed. 

In ablation study, we examine the effectiveness of modules in our architecture, lane representation ways (explicit point-wise \textit{vs.} implicit \textit{ELM}) and comparison between the mean square error (MSE) and the EIE loss on learning \textit{ELM}.

The main contributions can be concluded as follow: 
\begin{itemize}
\item We construct an end-to-end efficient geometric flexible lane detection framework ElasticLaneNet, which is composed of Encoder-Transformer-FPN pipeline. The sub-modules and feature fusion module are trained jointly to get \textit{ELM} output, and the training is guided by the EIE loss function.

\item A novel representation of lanes \textit{ELM} is proposed in Convolutional Neural Networks (CNN). \textit{ELM} contains complete lane information and is good at representing complex geometries in implicit form, giving a new perspective for lane modeling.

\item EIE loss is introduced for lane detection in our framework, which guides the neural network to approximate precise order, shape and position of diverse structure of lanes. Our energy-informed training strategy considers lanes structure as a whole, thus it well integrates the global context and the low level image features, enabling inference even under weak lane features.

\item The training strategy achieves state-of-the-art performance on the structurally diverse benchmark SDLane with competitive inference speed. The experimental results show that the proposed method is superior to the existing methods in the prediction of complex geometric lanes such as dense lanes, cross lanes, big turn lanes and Y-shaped lanes. F1 scores, Precision and Recall are 89.51, 91.61 and 87.50, respectively.
\end{itemize}

\section{Related Work}
\label{sec:relatedwork}

Nowadays, lane detection problems are primarily handled by the following categories of methods through neural network: segmentation-based, parameter-based, anchor-based, and row-wise methods.


\subsection{Segmentation-based methods}

Segmentation-based methods~\cite{pan2018spatial,hou2019learning,zheng2021resa,abualsaud2021laneaf,xu2020curvelane,neven2018towards,feng2023elastic} require an instance-level segmentation. After getting the pixel-wise discrimination, post-processing is needed to specify each lanes' coordinates. However, it is not efficient to regard the lanes as dense pixel masks~\cite{chougule2018reliable,qin2020ultra,liu2021condlanenet}. Performing pixel-wise classification without considering the global context of structured lanes and then dense post processing cause high computation cost as well as slow inference speed. Besides, the predicted lanes tend to be disconnected or missed when lane features are weak. 


\subsection{Parameter-based methods}
Parameter-based methods~\cite{tabelini2021polylanenet,feng2022rethinking,liu2021end} use explicit parametric functions to represent the lanes, such as polynomials~\cite{tabelini2021polylanenet}, Bezier curves~\cite{feng2022rethinking} or explicit functions based on a mathematical model~\cite{liu2021end}. The inference speed is fast because the parameters of the curves are relatively small. However, it is inflexible to represent the shape of lanes with specific functions. These approaches are very sensitive, \textit{i.e.} small error on high-order coefficient would cause a large deviation in the results. Therefore, explicit parameter-based methods are difficult to achieve high accuracy performance when the lanes geometries are diverse.


\subsection{Anchor-based methods}
Anchor-based methods~\cite{li2019line,tabelini2021keep,zheng2022clrnet,jin2022eigenlanes} firstly infer the region-of-interests (ROI) in the form of line anchors, and then calculate the offset from the anchors to construct the lanes. Line-CNN~\cite{li2019line} was the first work to propose the line-anchors method and LaneATT~\cite{tabelini2021keep} followed by adding global attention mechanism. CLRNet~\cite{zheng2022clrnet} is a SOTA method on TuSimple and CULane. However, these approaches rely on accurate prior anchors prediction, thus is not robust to complex topologies such as big turns, Y-shape or dense lanes. 
EigenLanes~\cite{jin2022eigenlanes} represents the complex lanes shape with the top rank Eigenlanes as anchors, and proposed a novel structure diversity dataset SDLane to evaluate their method. Our approach achieves higher score on this dataset in experiment.

\subsection{Row-wise methods}
The majority row-wise methods\cite{yoo2020end,qin2020ultra,qin2022ultra,liu2021condlanenet,jin2022eigenlanes} share some similarity to coarse grid multi-instance segmentation with higher efficiency. They output the coarse position in a down scale grid maps and obtain the x-coordinates of lanes according to coarse y samplings. Road shape information can be integrated into the lane models. However, these coarse confidence map may produce non-smooth portions or wrong lane locations when selecting points from sparse grid maps via post processing, especially when the lanes are high-curvature turns, Y-shaped, or parallel to the X-axis. 

Our method can be regarded as an efficient row-wise geometry-flexible approach with effective cross-row model informed by elastic interaction energy.

\section{Methodology}\label{sec:methodology}

In lane detection problem, a lane is usually described by its $x$-coordinates on the equidistant $y$-coordinate row, \text{i.e.}, the coordinates of the discretized lane are $\left\{\left(x_{i}, y_{i}\right)\right\}_{i=1}^M$, where $M$ is the number of sample points along $y$-direction.

In our ElasticLaneNet, given an input image $I \in R^{ H \times W \times 3}$, after down scaling into $I' \in R^{h\times w \times 3}$ and flowing through the network backbone, $N$ implicit function \textit{ELM}s with size $M\times w$, \text{i.e.} $O \in R^{N \times M\times w}$ (see Fig.~\ref{fig:ELM}) will be generated from the outputs of \textit{ELMM}, in which $N$ is the max number of lanes in a dataset; see Fig.~\ref{fig:networkarchitecture}. With the output of the \textit{ELM} $\Psi^k$'s,
the coordinates of each lane can be obtained from $\Psi^k=0$ along the $y$-direction; see Fig.~\ref{fig:ELM.image} and \ref{fig:ELM}. Because we rearrange the ground truth (GT) in order before training, when lanes number $n$ is smaller than $N$, the negative instances will be located at the end of outputs.

\subsection{Elastic Lane Map (\textit{ELM}) Representation}\label{subsec:ELM}

In our ElasticLaneNet, the lanes are represented as zero-level contour lines of \textit{ELM} that are evolving in the training process to identify the correct locations of the lanes.

\begin{figure}[t]
\centering

	\begin{minipage}[c]{0.3\linewidth}
		\centering
  
		\includegraphics[width=\textwidth]{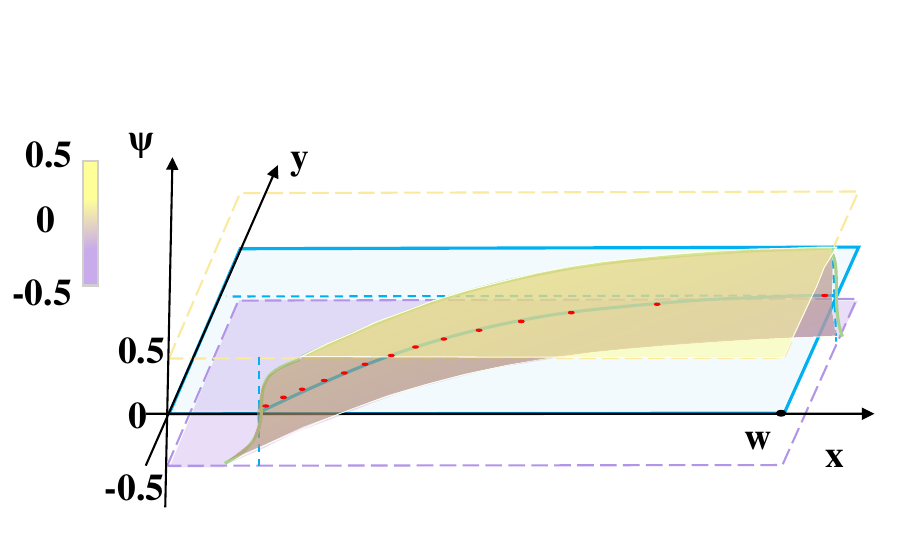}
		\caption{Elastic Lane Map, \textit{ELM} of one lane line.}
		\label{fig:ELM}
	\end{minipage} \hspace{3mm}
	\begin{minipage}[c]{0.27\linewidth}
		\centering
		\includegraphics[width=\textwidth]{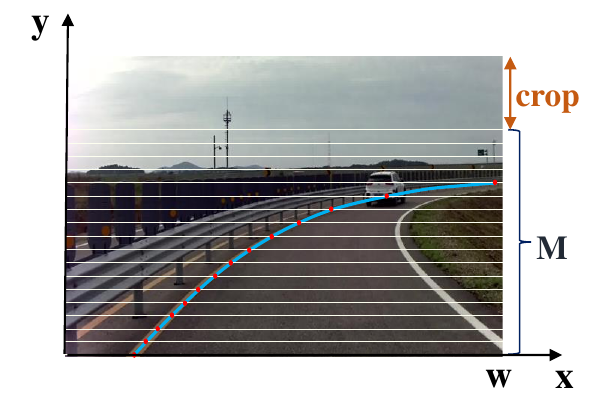}
		\caption{Row-wise lane points in an image.}
		\label{fig:ELM.image}
	\end{minipage} \hspace{3mm}
        \begin{minipage}[c]{0.27\linewidth}
		\centering
		\includegraphics[width=\textwidth]{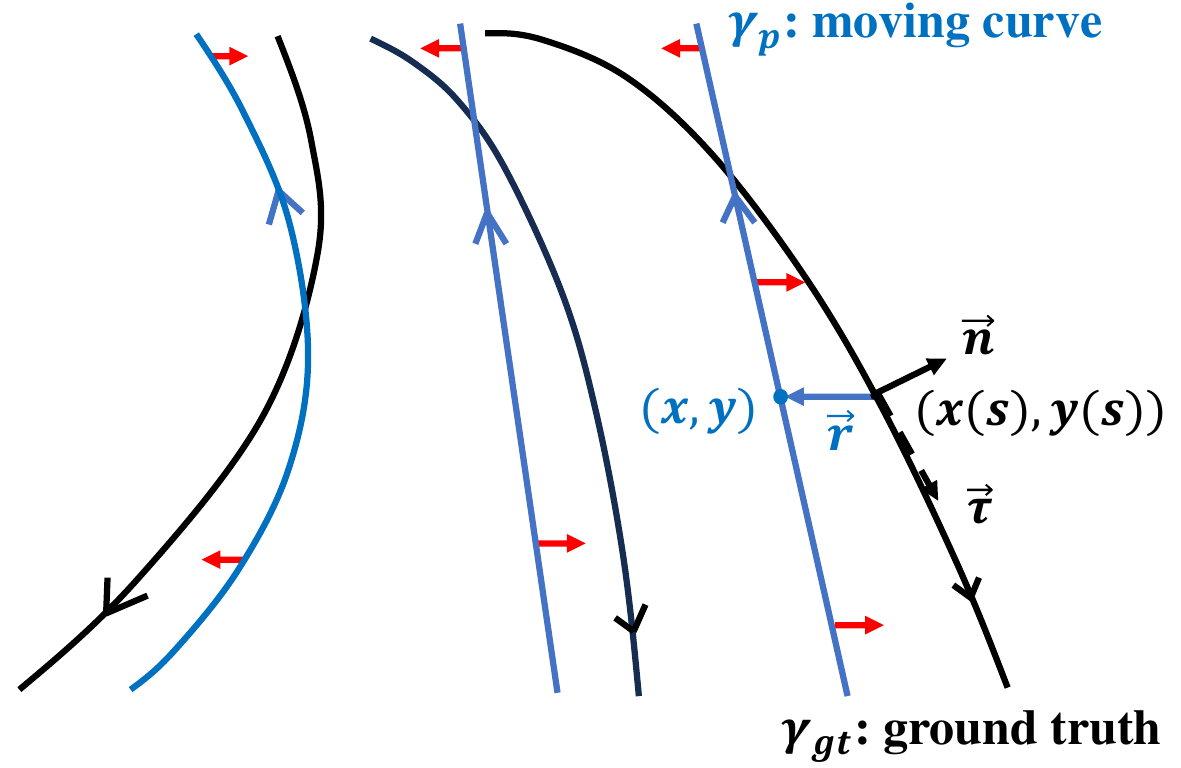}
		\caption{Lanes moving informed by the EIE loss.}
		\label{fig:movinglanes}
	\end{minipage}
\end{figure}

In our method, a lane is implicitly represented by \textit{ELM}: $\Psi=H_\sigma(\phi)-0.5$ and the lane is the curve given by $\Psi=0$.
Here $\phi$ is the level set function \cite{levelset} of the lane: $\phi(x, y)= -d(x, y), \text { if }(x, y) \in \gamma^-; \phi(x, y)= 0, \text { if }(x, y) \in \gamma; \phi(x, y)= d(x, y), \text { if }(x, y) \in \gamma^+, $
where $\gamma^-$ and $\gamma^+$ respectively are the left and right sides of the lane curve $\gamma$, and $d(x,y)$ is the $x$-distance to the lane curve with the same $y$-coordinate; and 
$ H_\sigma(\cdot)$ is the smoothed Heaviside function: $H_{\sigma}(\phi)= 0,  \text { if } \phi \leqslant-\sigma; H_{\sigma}(\phi)= \frac{1}{2}(1 +\frac{\phi}{\sigma}), \text { if }-\sigma<\phi<\sigma; H_{\sigma}(\phi)= 1, \text { if } \phi \geqslant \sigma. $
Figure~\ref{fig:ELM} is an illustration plot of \textit{ELM}. When there are at most $N$ lanes in an image, $N$ \textit{ELM}s $\Psi^k, k=1,2,...,N$, are employed, with the curve $\Psi^k=0$ representing the $k$th lane; see the illustration of the \textit{ELM} in the architecture of our network in Fig.~\ref{fig:networkarchitecture}. The smoothed Heaviside function $H_\sigma(\cdot)$ can be approximated by the Sigmoid or Softmax function in the network.



The \textit{ELM}s are evolved under the guided via the EIE loss in the training process. Lanes prediction and ground truth generate a long-range interaction, which attracts the curves lied on the \textit{ELM}s to the ground truth; see ~\ref{subsec:EIE}.







\subsection{Elastic Interaction Energy (EIE) Loss}\label{subsec:EIE}

In lane detection problem, the predicted lane and the ground truth can be liken to two defected lines in crystal dislocation \cite{xiang2006active}. 
The elastic interaction energy of a system of a set of curves $\gamma$ is
$
E =\frac{1}{8 \pi} \int_{\gamma} \int_{\gamma^{\prime}} \frac{d \boldsymbol{l} \cdot d \boldsymbol{l}^{\prime}}{r},$
where vector $d \boldsymbol{l}$ represents line element on curves $\gamma$ with tangent direction $\boldsymbol{\tau}$, \text{i.e.} $d\boldsymbol{l}=\boldsymbol{\tau} d l$, $\gamma^{\prime}$ denotes the curves with another parameter, and the $r$ is Euclidean distance between a point $(x,y)$ on $\gamma$ and a point $\left(x^{\prime}, y^{\prime}\right)$ on $\gamma^{\prime}$, \text{i.e.} $r=\sqrt{\left(x-x^{\prime}\right)^2+\left(y-y^{\prime}\right)^2}$. 


In our ElasticLaneNet, we define $\gamma=\gamma_{gt}\cup\gamma_p$ for a lane, where $\gamma_{gt}$ and $\gamma_p$, 
are the ground truth and the prediction of the lane, respectively.
The above elastic interaction energy of this system is
\begin{equation}\label{eq.eie2}
\begin{aligned}
E & =\frac{1}{8 \pi} \int_{\gamma_{gt}\cup\gamma_{gt}^{\prime}} \int_{\gamma_{p}\cup\gamma_{p}^{\prime}} \frac{(d \boldsymbol{l}_{\mathbf{gt}} + d \boldsymbol{l}_{\mathbf{p}}) \cdot (d \boldsymbol{l}_{\mathbf{gt}}^{\prime} + d \boldsymbol{l}_{\mathbf{p}}^{\prime})}{r} \\
&=\frac{1}{8 \pi} \int_{\gamma_{gt}} \int_{\gamma_{gt}^{\prime}} \frac{d \boldsymbol{l}_{\mathbf{gt}} \cdot d \boldsymbol{l}_{\mathbf{gt}}^{\prime}}{r}+\frac{1}{8 \pi} \int_{\gamma_p} \int_{\gamma_p^{\prime}} \frac{d \boldsymbol{l}_{\mathbf{p}} \cdot d \boldsymbol{l}_{\mathbf{p}}^{\prime}}{r} +\frac{1}{4 \pi} \int_{\gamma_{gt}} \int_{\gamma_{p}} \frac{d \boldsymbol{l}_{\mathbf{gt}} \cdot d \boldsymbol{l}_{\mathbf{p}}}{r}
\end{aligned}
\end{equation}
The energy in Eq.~\ref{eq.eie2} can be written as $E=E_s+E_i$, where the self-energy $E_s$ consists of the first two terms, \text{i.e.}, the self-energies of the ground truth curve $\gamma_{gt}$ and the prediction curve  $\gamma_p$, respectively, while the third term is the interaction energy $E_i$ between the two  curves $\gamma_p$ and $\gamma_{gt}$.

An important property of this elastic interaction energy is that when the prediction $\gamma_p$ and the ground truth $\gamma_{gt}$ have opposite orientations, the $\gamma_p$ will be attracted to $\gamma_{gt}$ to minimize the total energy in Eq.~\ref{eq.eie2}, and such attractive interaction is long-range \cite{xiang2006active,lan2020elastic}. See Fig.~\ref{fig:movinglanes} for an illustration. This can also be intuitively seen by the fact that if the prediction $\gamma_p$ and the ground truth $\gamma_{gt}$ coincide but have opposite orientations, $\gamma=\gamma_{gt}\cup\gamma_p$ will be completely annihilated, \text{i.e.} $d \boldsymbol{l}_{\mathbf{p}}=-d \boldsymbol{l}_{\mathbf{gt}}$, giving the minimum value $0$ for the elastic interaction energy in Eq.~\ref{eq.eie2}.
Besides, the self-energy of the prediction $\gamma_p$ tends to make it smooth, because a non-smooth curve is longer and has a larger self-energy.

   


In our ElasticLaneNet, under the \textit{ELM} representation presented in the previous subsection, we use the 
following elastic interaction energy  (EIE) Loss:
\begin{equation}\label{eq.leie}
\mathcal{L}_{eie}(\Psi_p,G_t) =\frac{1}{8 \pi} \int_{\mathbf{R}^2}d x^{\prime} d y^{\prime}\int_{\mathbf{R}^2} \frac{\nabla\left(G_t-\alpha \Psi_p\right)(x, y) \cdot \nabla\left(G_t-\alpha \Psi_p\right)\left(x^{\prime}, y^{\prime}\right)}{r} d x d y.
\end{equation}
where $G_t=H_\sigma(\phi_{gt})-0.5$ for the ground truth of the lane and $\Psi_p=H_\sigma(\phi_p)-0.5$ for the implicit function \textit{ELM} for the prediction, and $\alpha$ is a parameter. Here we have used $\nabla H(\phi_{gt}) = -\delta (\gamma_{gt})\boldsymbol{n}$ and $\nabla H(\phi_p) = -\delta (\gamma_{p})\boldsymbol{n}$ in Eq.~\ref{eq.eie2} to obtain this loss, where $\delta (\gamma)$ is the Delta function of the curve $\gamma$ and $\boldsymbol{n}$ is its normal direction.
Note that the "$-$" sign in $G_t-\alpha \Psi_p$ in Eq.~\ref{eq.leie} ensures the opposite orientations of $\gamma_{gt}$ and $\gamma_p$ represented by $G_t$ and $\Psi_p$, respectively. 

In order to reduce computation cost ($O(N^2)$ in direct integration), we apply fast Fourier Transform (FFT) in the implementation to efficiently calculate the gradient of this EIE loss \cite{lan2020elastic} and the computation cost $O(N\log N)$, which is
\begin{equation}\label{eq.veie.fft}
\frac{\partial \mathcal{L}_{eie}}{\partial \phi}=\mathcal{F}^{-1}\left(\frac{\sqrt{m^2+n^2}}{2} d_{m n}\right),
\end{equation}
where $m$ and $n$ here are the frequencies in Fourier space, $d_{m n}$ is the Fast Fourier Transform of $G_t-\alpha \Psi_p$,
and $\mathcal{F}^{-1}$ is the inverse Fourier transform.

\subsection{Network Architecture}
\subsubsection{Main Stream of Network Structure}
The main stream of our ElasticLaneNet is constructed by backbone with encoder as ResNet34\cite{he2016deep} and decoder as feature pyramid network (FPN)\cite{lin2017feature}, and the \textit{ELM} outputs from the backbone. In addition, the transformer bottleneck (TB)\cite{vaswani2017attention,liu2021condlanenet} and three sub-branches are designed to improve the performance, including auxiliary feature refinement (AFR) module, lane existence classification sub-network (CSN) and lane range sub-network (RSN). The details of the network structure can be seen in Fig.~\ref{fig:networkarchitecture}. The version with TB is named ElasticLaneNet$^\text{T}$.


\subsubsection{Auxiliary Feature Refinement.}
Inspired by cross layer refinement\cite{zheng2022clrnet}, we design an AFR module in FPN to gather cross layer features. A feature fusion scheme (FF) are performed via concatenating $P_1$ layer with $F_2$, $F_3$ (upsampled from $P_2$ and $P_3$ layers, respectively) in AFR for the final implicit lanes prediction. 

Besides, in AFR module, an auxiliary deep supervision\cite{wang2015training} scheme is applied only in training process for features integration, through the auxiliary EIE loss $\mathcal{L}_{aux}$, \text{i.e.} $ \mathcal{L}_{aux} = \lambda_{1}*\mathcal{L}_{eie}(\Psi_{p_2},G_t) + \lambda_{2}*\mathcal{L}_{eie}(\Psi_{p_3},G_t), $
where $\Psi_{p_2}$ and $\Psi_{p_3}$ are produced from the predictions of $P_2$ and $P_3$ layers in FPN, see Fig.~\ref{fig:networkarchitecture}. 

\subsubsection{Lane Range and Existence.}
We add the CSN and RSN at the bottle neck layer for jointly learning the existence and length of each lane in \textit{ELM}. 

The classification branch CSN is designed for no-lane situation, trained by a lane existence loss $\mathcal{L}_{exist}$ via a commonly used Focal Loss \cite{lin2017focal,liu2021condlanenet,duan2019centernet,law2018cornernet}. 

The end of the lane on the predicted \textit{ELM} occasionally has a departure point. Although these points can be eliminated in post-process, an end-to-end approach is more efficient. Inspired by ~\cite{qin2020ultra,liu2021condlanenet}, the binary-cross-entropy loss $\mathcal{L}_{range}$ is applied to learn whether a certain lane marking exists on each row $y_i$, thus obtaining the lane range: 
$
\mathcal{L}_{range}=-\frac{1}{M} \sum_{i=1}^M y_i \cdot \log \left(p\left(y_i\right)\right)+\left(1-y_i\right) \cdot \log \left(1-p\left(y_i\right)\right),$
where $p\left(y_i\right)$ is the predicted positive probability on $y_i$. 

The portions of the lanes that are out of range or classified as having no lines will be discarded, resulting in the final \textit{ELM}.


\subsubsection{Total Loss.}

ElasticLaneNet is a multi-task learning strategy, in which the total loss function is the weighted sum of the elastic loss $\mathcal{L}_{eie}$, the existence loss $\mathcal{L}_{exist}$, the range loss $\mathcal{L}_{range}$ and the auxiliary loss $\mathcal{L}_{aux}$. 
The total loss function is:
\begin{equation}
\mathcal{L}_{total} = \lambda_{eie}\mathcal{L}_{eie} + \lambda_{a}\mathcal{L}_{aux} + \lambda_{r}\mathcal{L}_{range} + \lambda_{e}\mathcal{L}_{exist}
\end{equation}

\section{Experiments}\label{sec:experiments}

\subsection{Datasets and Evaluation Metrics}

\subsubsection{Datasets}
In this paper, three datasets are applied for evaluation: SDLane~\cite{jin2022eigenlanes}, CULane~\cite{pan2018spatial} and TuSimple~\cite{tusimple2017benchmark}. SDLane~\cite{jin2022eigenlanes} is a newly proposed dataset with up to 7 lanes and a variety of high complexity of lane structures, including intersections, Y-shape forks, confluence roads, dense lanes, widely distributed curves, etc. In CULane~\cite{pan2018spatial}, the driving scenes are divided into nine categories with 4 lanes at most, which include several complex scenes such as crowded, night, shadow and dazzling images. It is noticed that in SDLane, any drivable areas are considered to have lanes, \textit{e.g.}, a crossroad that allows vehicles to turn left and go straight is considered as two lanes with four lane lines, while CULane regards the crossroad as no lane. TuSimple~\cite{tusimple2017benchmark} is a dataset of highways in good weather with maximum 5 lanes. The proportion of curves in SDLane is grater than 90\%, while it is about 2\% in CULane and 30\% in TuSimple.

\subsubsection{Evaluation Metrics}

In this paper, the official metrics for CULane ~\cite{pan2018spatial} are applied on evaluation of CULane and SDLane, \textit{i.e.} F1-measure ($F1$) based on Intersection over Union (IoU), $Precision$ and $Recall$, which are
$
F 1=2 \cdot\left(\frac{precision  \cdot  recall }{precision +  recall}\right),
$
$Precision=\frac{TP}{TP+FP}$, $Recall=\frac{TP}{TP+FN}$. Here $TP$ is the correctly predicted lane points, $FP$ is the number of false positives, and $FN$ is number of false negatives.

The official metrics of the performance on TuSimple~\cite{tusimple2017benchmark} are accuracy (Acc), false positive (FP) and false negetive (FN). $ \text { Acc }=\frac{N_{p r e d}}{N_{g t}} $,
where $N_{p r e d}$ is the number of correct predicted lane points and $N_{g t}$ is the number of ground-truth lane points. In addition, F1-measure is also applied in our TuSimple experiment.

\subsection{Implementation Details}\label{sec:implementation}

Our experiments are implemented on a server of NVIDIA RTX 4000. The parameters $\lambda_{eie}$, $\lambda_{aux}$,  $\lambda_{range}$ and  $\lambda_{exist}$ in the loss functions are 1.0, 1.0, 0.1, 0.2. In $\mathcal{L}_{aux}$, both of the parameters $\lambda_1$ and $\lambda_2$ are set 0.3 when without feature fusion in AFR, and set 0 otherwise. The hyper-parameter $\alpha$ in the EIE loss is set as 0.5. $\sigma$ is the parameter in the Heaviside Function, which is set according to the number of sample $M$, \textit{i.e.} $M=36$ with $\sigma=3$ for SDLane and TuSimple, $M=18$ with $\sigma=5$ for CULane.


\subsection{Comparison with State-of-the-Art Methods}

\subsubsection{Comparison on SDLane.}\label{subsec.compare}

\begin{figure}[ht]
  \centering
   \includegraphics[width=1.0\textwidth]{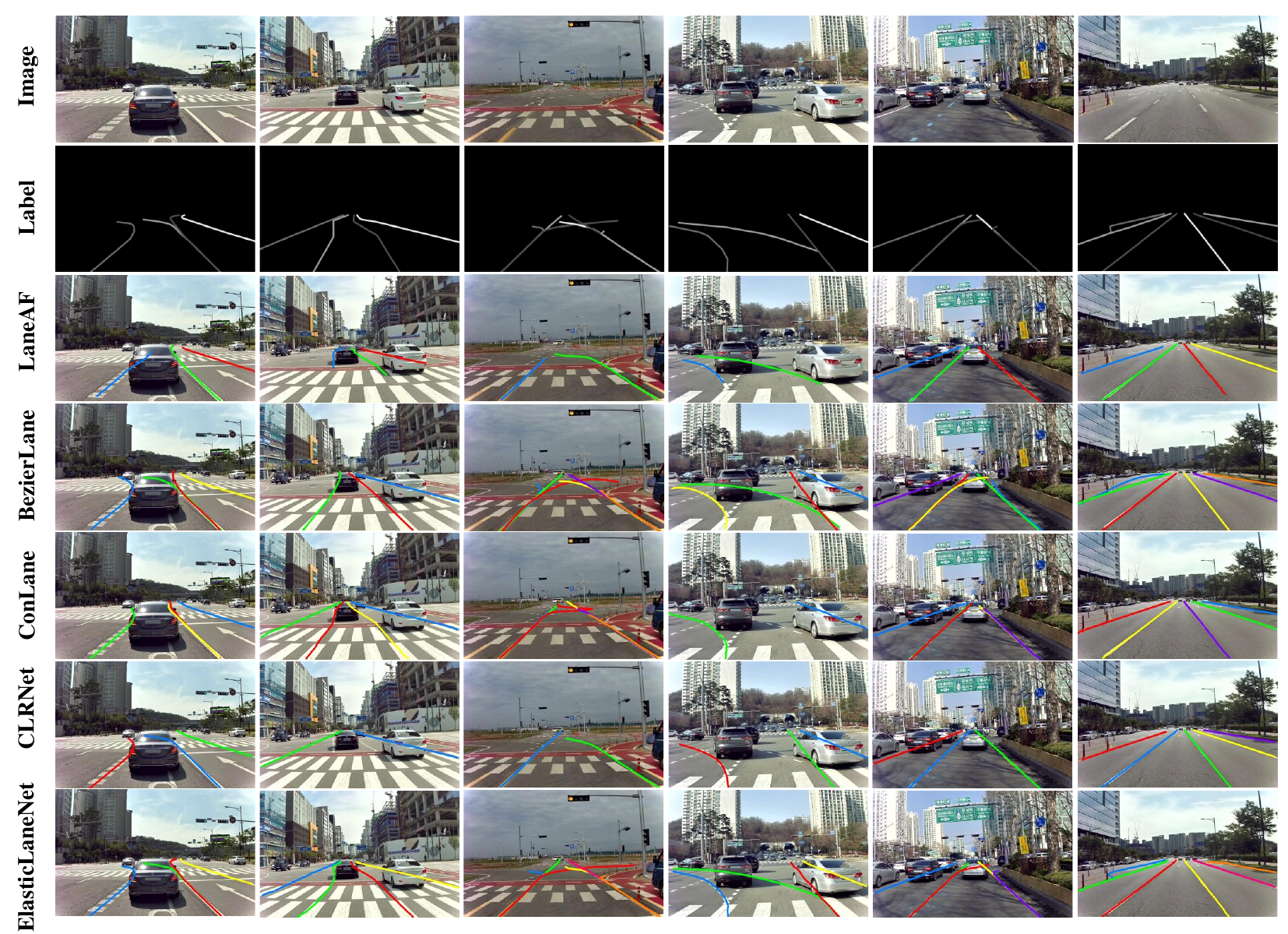}
   \caption{Comparison of the prediction results on challenging cases in SDLane using ElasticLaneNet and other state-of-the-art lane detection meodels. Different colors represent different instances.}
   \label{fig:sdlaneresults}
\end{figure}

Table~\ref{tab.sd} and Fig.~\ref{fig:sdlaneresults} show the experiments performed on structurally diverse lane data SDLane~\cite{jin2022eigenlanes} by our ElasticLaneNet and other models. We re-trained the most recently SOTA lane detection models of each type to convergence, and record the results in Tab.~\ref{tab.sd}. Our ElasticLaneNet outperforms all other methods on $F1$ score and $Recall$. High $Recall$ indicates less likely to miss prediction, reflecting the strong inference ability on various difficult positive samples. High $Precision$ means small portion of redundant predictions. $F1$ is a balance of these two metrics. The results show the strong capability of our ElasticLaneNet on representing and detecting structure complex cases.

\begin{table}[h]\small
\caption{Comparison of our ElasticLaneNet with state-of-the-art lane detection methods on SDLane. The methods marked with '\dag' are implementation results in~\cite{jin2022eigenlanes}.}\label{tab.sd}
\setlength{\abovecaptionskip}{2mm}
\setlength{\belowcaptionskip}{-2mm}
    \centering
    \resizebox{0.50\textwidth}{!}{
    \begin{tabular}{@{}l|ccc|c@{}}
    \toprule
    Method &F1 & Precision & Recall & Type \\
    \midrule
    RESA$^\dag$~\cite{zheng2021resa}  &77.09 &82.35 &72.46 & \multirow{2}{*}{Seg.-based}\\
    LaneAF~\cite{abualsaud2021laneaf}      &74.17 &87.15 &64.55 &    \\
    \midrule
    
    BazierLane~\cite{feng2022rethinking} &77.13   &77.84  &76.43 &Para.-based \\
    \midrule
    LaneATT$^\dag$~\cite{tabelini2021keep} &73.49   &85.78  &64.28 &\multirow{3}{*}{Anchor-based}\\
    EigenLanes$^\dag$~\cite{jin2022eigenlanes} &80.47   &86.04  &75.58 & \\
    CLRNet~\cite{zheng2022clrnet} &85.04  &\bf{94.16}   &77.53 &  \\
    \midrule
    CondLane$^\dag$~\cite{liu2021condlanenet} &75.97 &87.59  &67.08 & \multirow{3}{*}{Row-wise} \\
    CondLane$^\text{R}$~\cite{liu2021condlanenet} &87.02 &87.73 &86.32 &  \\
    UFLDv2~\cite{qin2022ultra}  &65.41   &80.21  &55.21 &        \\
    \midrule
    ElasticLaneNet  &\underline{89.51}  &91.61  &\underline{87.50} &\multirow{2}{*}{Ours}\\
    ElasticLaneNet$^\text{T}$ & \bf{90.88} & 92.87 & \bf{88.97} & \\

    \bottomrule
    \end{tabular}
    }
\end{table}

From the last row of  Fig.~\ref{fig:sdlaneresults}, ElasticLaneNet has successfully identified a variety of challenging samples, such as the left-turning line and straight lines blocked by cars at intersections, winding roads, six lines at three-way junctions, big bends with no line on the ground, newly added Y-shaped lanes, dense lanes, etc. In addition, the lanes predicted by ElasticLaneNet are smooth.

Figure~\ref{fig:sdlaneresults}, rows 3-7 are the comparison between our ElasticLaneNet with current SOTA and different types of popular lane detection models on SDLane. The segmentation-based LaneAF~\cite{abualsaud2021laneaf} misses or predicts incompletely when targets are blocked and the lanes are not obvious. It reflects that this type of methods heavily rely on low level image features, and cannot handle weak lane feature situation well. The predictions of parametric-based BezierLaneNet~\cite{feng2022rethinking} deviate significantly from the ground truth when the lane shapes are diversified, although they can explore many challenging instances. CondLaneNet~\cite{liu2021condlanenet} is an effective row-wise method. It also designs an auxiliary module named RIM to handle fork lanes. However, when facing large forks or weak feature, it sometimes cannot give precise shape and location of the branch lanes. Although the current state-of-the-art CLRNet~\cite{zheng2022clrnet} (anchor-based method) achieve the highest $Precision$ in Tab.~\ref{tab.sd}, \textit{i.e.} the least redundant prediction, it tends to miss dense lanes, Y-shaped lanes and big bends, which are crucial information for an autonomous vehicle. The prior anchors seems not flexible enough to represent structurally diverse lanes in these examples.

\begin{table}[htbp]
\caption{FPS comparison on RTX 4000 (SDLane). }\label{tab:fps}
\centering
\setlength{\abovecaptionskip}{4mm}
\setlength{\belowcaptionskip}{-5mm}
    \resizebox{0.70\textwidth}{!}{
    \begin{tabular}{*{7}{c}}
        \toprule
        Method & LaneAF  & CLRNet & CondLane  &ElasticLane &ElasticLane$^\text{T}$ &  BezierLane\\
        \midrule
        Tpye & Seg. &Anchor.  &Row. & \multicolumn{2}{c}{Ours} & Para. \\
        \midrule
        FPS & 11.29 & 59.26 & 61.95 & \underline{75.42} & \underline{66.62} &\textbf{109.37} \\
        \bottomrule
    \end{tabular}
    }
    
\end{table}

Table~\ref{tab:fps} displays the inference speed comparison on Frames Per Second (FPS) between our ElasticLaneNet and other methods we re-implemented on our server RTX 4000. Except for the parameter-based method BezierLaneNet~\cite{feng2022rethinking} that achieving the fastest speed (109.37 FPS), our method ranks on the 2nd with 75.42 FPS, even with the TB module (66.62 FPS). However, the prediction of BezierLaneNet seems not quite accurate; see Tab.~\ref{tab.sd} and Fig.~\ref{fig:sdlaneresults}. The segmentation-based method LaneAF~\cite{abualsaud2021laneaf} requires very time-consuming post process, anchor-based CLRNet~\cite{zheng2022clrnet} needs traverse Non-Maximum Suppress (NMS) to select the best prediction, and row-wise method CondLaneNet~\cite{liu2021condlanenet} takes time to integrate the network outputs, \textit{i.e.} confidence maps, offset maps and start points maps in post process. Our approach well balance the effectiveness and efficiency. 


\begin{table}[b]\small

\caption{Comparison of our methods with other popular models under relatively light weight backbone, ResNet34 in majority, on TuSimple. }\label{tab.tu}
    \centering
    \resizebox{0.60\textwidth}{!}{
    \begin{tabular}{l|ccccc}
    \toprule
    Method  & Backbone &F1 &Acc & FP & FN \\
    \midrule  
    LaneAF~\cite{abualsaud2021laneaf}  & ERFNet &96.49 &95.62  &0.0280  &0.0418 \\
    
    UFLD~\cite{qin2020ultra} & ResNet34 &- &96.06  &0.1891  &0.0375\\

    UFLDv2~\cite{qin2022ultra} & ResNet34 &95.56 &96.22  &0.0318  &0.0437\\
    LaneATT~\cite{tabelini2021keep} & ResNet34 &96.77 &95.63 &0.0353  &\bf{0.0292} \\
    CondLane~\cite{liu2021condlanenet} & ResNet34 &96.98 &95.37   &\bf{0.0220}   &0.0382 \\
    BazierLane~\cite{feng2022rethinking} & ResNet34 &- &95.65  &0.0510  &0.0390\\
    EigenLanes~\cite{jin2022eigenlanes} & ResNet50 &-   &95.62   &0.0320   &0.0399 \\
    
    LSTR~\cite{liu2021end} & Transformer &96.68 &96.18  &0.0291  &0.0338\\
    
    \midrule
    ElasticLaneNet & ResNet34 & 96.36 &95.90 &0.0280  &0.0430 \\
    ElasticLaneNet$^\text{T}$ &ResNet34 &\bf{97.05} &\bf{96.48} &0.0274  &0.0315 \\
    \bottomrule
    \end{tabular}
    }

\end{table}


\subsubsection{Comparison on TuSimple.}
\begin{figure}[ht]
  \centering
   \includegraphics[width=0.7\textwidth]{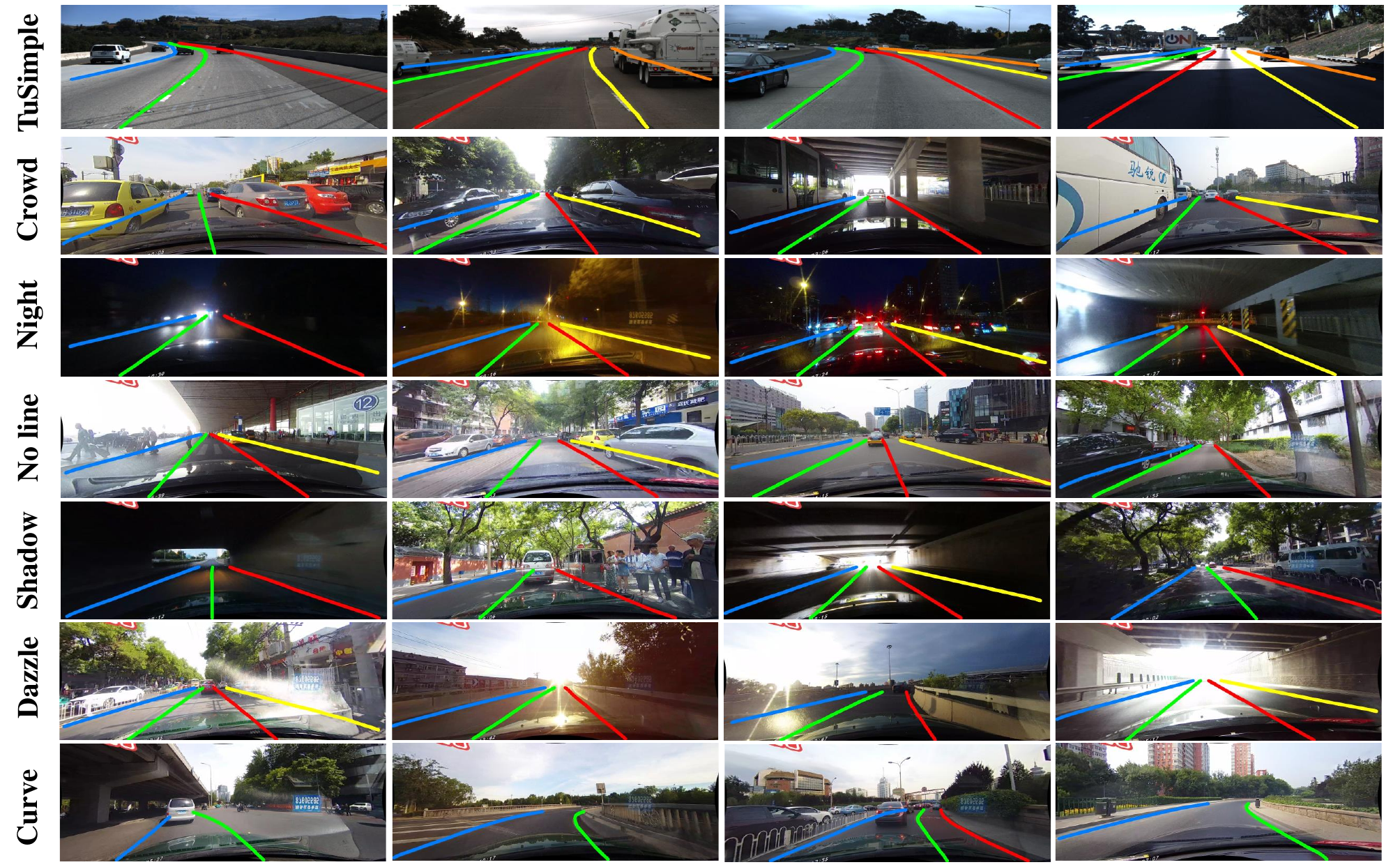}
   \caption{Prediction results of challenging cases on TuSimple (the first row) and CULane (the remaining rows) via ElasticLaneNet.}
   \label{fig:culaneresults}

\end{figure}

TuSimple \cite{tusimple2017benchmark} is an early lane detection dataset that mainly covers driving scenarios on highways with good lighting conditions. A comparison of ElasticLaneNet with most recent lane detection models (ResNet34 backbone versions in majority) shows that our approach is competitive and outperforms most of the existing models, particularly in fitting large curvature curves; see Tab.~\ref{tab.tu}. Our results are not far away from the current SOTA to our knowledge (both metrics are presented at the same time, \textit{i.e.} 97.82 F1 and 96.9\% Accuracy according to~\cite{zheng2022clrnet}). Some excellent results of challenging cases in TuSimple are presented in the first row of  Fig.~\ref{fig:culaneresults}, \textit{e.g.} large curve, dense or side lanes, and weak features under occlusion or shadow.


\subsubsection{Comparison on CULane.}
Experimental results show that ElasticLaneNet also performs very well on CULane~\cite{pan2018spatial}, a commonly used dataset with diverse scenarios. In many complex scenarios, we have accurate predictions; see Tab. Table~\ref{tab.cu} and Fig.~\ref{fig:culaneresults}. Since most of CULane's lane lines are straight lines, 18 sample points can achieve good results with an efficient inference speed in ElasticLaneNet. The FPS of inference tested in RTX 4000 is 105/90 (without/with TB module), which is much faster than 36 sampling points in Tab.~\ref{tab:fps}. Figure~\ref{fig:culaneresults} shows some good predictions of ElasticLaneNet on challenging cases in CULane. Because of the long range nature of our method, we can handle weak features, \textit{e.g.} completely occluded by car or dazzle light, totally dark night with no line, traffic scene with lots of people and cars, etc. However, the current SOTA methods are anchor-based approach to our knowledge~\cite{zheng2022clrnet,honda2024clrernet}.This might because the lane structures are mostly straight lanes in CULane, the advantage of our model is not significant.


\begin{table*}[h]\tiny
\caption{Comparison of the total/category F1 score (threshold IoU = 0.5) on CULane using ElasticLaneNet and other recent methods. For “Cross” category , only $FP$ are shown. (\textit{ELM}-MSE means applying MSE instead of 
the EIE loss in our framework. Underlined are those with performance improvements of more than 3 F1 score.)}\label{tab.cu}
\resizebox{\textwidth}{18mm}{
\begin{tabular*}{\hsize}{@{}@{\extracolsep{\fill}}l|cccccccccc@{}}
\toprule
Method &Total &Normal  &Crowded  &Dazzle  &Shadow  &No line   &Arrow   &Curve  &Cross  &Night  \\
\midrule  
SAD~\cite{hou2019learning} &71.80 &90.70   &70.00  &59.90  &67.00  &43.50  &84.40  &65.70  &2052  &66.30 \\
UFLD~\cite{qin2020ultra} &72.30 &90.70   &70.20  &59.50  &69.30  &44.40  &85.70  &69.70  &2037  &66.70 \\
LaneAF~\cite{abualsaud2021laneaf} &75.63   &91.10   &73.32   &69.71   &75.81   &\bf{50.62}   &86.86   &65.02   &1844   &70.90 \\
RESA-50~\cite{zheng2021resa} &75.30  &92.10   &73.10   &69.20   &72.80   &47.70   &88.30   &\bf{70.30}   &1503   &69.90  \\
BazierLane~\cite{feng2022rethinking} &75.57 &91.59   &73.20  &69.20  &\bf{76.74}  &48.05  &87.16  &62.45  &\bf{888}  &69.90  \\

UFLDv2~\cite{qin2022ultra} &75.9 &\bf{92.5}   &\bf{74.9}  &65.7  &75.3  &49.0  &88.5  &69.70  &1864  &70.6 \\
\midrule
\textit{ELM}-MSE &72.88  &90.55  &71.38   &64.31  &68.62  &44.00   &87.05   &65.95  &1744   &66.99 \\
\midrule
ElasticLaneNet &75.74 &91.95  &73.71  &\underline{\bf{72.34}}  &\underline{74.54}   &\underline{47.51}   &88.44   &\underline{69.12}  &1687   &\underline{70.66}  \\
ElasticLaneNet$^\text{T}$ &\bf{75.98}  &92.48  &74.29  &69.43  &73.72 &48.63   &\bf{88.95}   &68.66  &1578   &\bf{71.12}  \\

\bottomrule
\end{tabular*}
}

\end{table*}


\subsection{Ablation Study}

\subsubsection{Ablation Study on Network Sturcture}
An overall ablation study of our ElasticLaneNet architecture is summarized in Tab.~\ref{tab.ablation}. The baseline model on the first row only apply MSE on the \textit{ELM} without trained with CSN and RSN (CRSN for simplicity). From the 2nd row in Tab.~\ref{tab.ablation}, the EIE loss leads to a significant improvement on structure diverse lanes in SDLane. 

CRSN is designed to replace post processing (Post.P) for \textit{ELM} refinement. According to Tab.~\ref{tab.ablation}, applying CRSN instead of Post.P improves the F1 score as well as Precision, while causing a trade off on Recall. Recall shows the model's learning ability for positive samples. In the experimental comparison, we find that although the CRSN modules can improve the accuracy of predicting correct positive samples, the network loses part of the learning ability on some challenging scenes, \textit{e.g.} dense lanes on both sides of the image. Such results indicate that the prediction becomes more conservative with CRSN. Note that the Post.P scheme here is simply removing points that have been shifted by more than 50 because the implicit function \textit{ELM} sometimes descends along the $y$-axis as well. 

In Auxiliary Feature Refinement module, the cross layers FF module increases the overall performance as shown in Tab.~\ref{tab.ablation}. Also, in-order to look deeper into the role of features in latent space, we insert TB module, leading to significant improvement as shown in Tab.~\ref{tab.sd},~\ref{tab.tu} and~\ref{tab.ablation}. This infers that self-attention mechanism improves the accuracy of CRSN prediction, which is of great help to shape the ELM.


\begin{table}[t]\small
\caption{Ablation study of ElasticLaneNet on SDLane. The MSE loss is applied instead of the EIE loss as the baseline (first row).}\label{tab.ablation}
    \centering
    \resizebox{0.50\textwidth}{!}{
    \begin{tabular}{@{}ccccccccc@{}}
    \toprule
    Post.P &TB &FF &CRSN   &EIE loss & F1 & Recall & Precision & FPS\\
    \midrule
    $\checkmark$ & & & & & 79.49 & 79.59 & 79.38 & -\\
    $\checkmark$ & & & &$\checkmark$  & 86.34 & 86.92 & 85.77 &- \\
    & & &$\checkmark$ & $\checkmark$ & 87.13 & 84.54 & 89.87 &79.41 \\
    &  &$\checkmark$ &$\checkmark$  & $\checkmark$ & \underline{89.51} & \underline{87.50} & \underline{91.61} &75.42 \\
    & $\checkmark$  &$\checkmark$ &$\checkmark$ & $\checkmark$ & \bf{90.88} & \bf{88.97} & \bf{92.87} & 66.62 \\
    \bottomrule
    \end{tabular}
    }
\end{table}


\subsubsection{Ablation Study on Lane Representation}\label{subsubsec.LR}
In order to explore the most suitable lane representation, we examine a variant of the detection head instead of \textit{ELMM}, which is an explicit point-wise sampling head without FPN up-sampling. Given an input image $I \in R^{H \times W \times 3}$, the output from network is $O' \in R^{N \times M}$. The explicit network mapping trained by minimizing EIE loss is $f: I \mapsto O'$, where $O'=\left\{x_{1}^{\left(k\right)},x_{2}^{\left(k\right)},\ldots,x_{M}^{\left(k\right)}\right\}_{k=1}^N$, names ElasticLaneNet$^{pw}$ here. The advantages of ElasticLaneNet$^{pw}$ is having smaller model parameters and faster inference speed. Table~\ref{tab.pointwise} shows this prediction results of the variant ElasticLaneNet$^{pw}$. Compared with the results of ElasticLaneNet with \textit{ELMM} head in Tab.\ref{tab.sd}, the explicit point-wise approach fails on predicting geometry-complicated lanes in SDLane, with a dramaticlly degradation of -32.98\% comparing 89.51 F1 score. Also, in the experiment, ElasticLaneNet$^{pw}$ is not easy to converge to global minimum of energy function EIE, thus we prefer implicit \textit{ELM} in this study. Notably, the results of the ElasticLaneNet$^{pw}$ experimented on CULane only decrease slightly, suggesting that the explicit model has the potential to detect simple structural lanes at faster speed.

\subsubsection{Ablation Study on EIE Loss}
In order to illustrate the effectiveness of the EIE loss, we apply another classical loss function in supervised learning, the MSE on the implicit functions \textit{ELM} studying; see Tab.~\ref{tab.cu} and \ref{tab.losscompare}. In scene diverse data CULane and structure complex data SDLane, the EIE loss is much more effective than the MSE. In the experiment on CULane, the EIE loss obtains higher F1 score than MSE in all categories, among which the categories of dazzle, shadow, no line, curve, night are 3 points higher. In the SDLane experiment, the EIE loss outperform the MSE in all evaluation metrics with more than 6 points higher. In the TuSimple experiment trained with the EIE loss, the FP and FN are smaller, and accuracy higher. This ablation study shows the effectiveness of EIE-informed strategy for solving weak vision features image and complex-geometric lanes. 

We also examine explicit MSE-informed training. However, without any anchor or confidence map, explicit experiments with MSE loss fail due to training collapse, while the performance of ElasticLaneNet$^{pw}$ is acceptable.


\begin{minipage}[h]{0.48\textwidth}\
\makeatletter\def\@captype{table}
\caption{Performance via point-wise variant ElasticLaneNet$^{pw}$.}\label{tab.pointwise}
\centering
\resizebox{0.9\textwidth}{!}{
    \begin{tabular}{@{}c|cccc@{}}
    \toprule
    Metric &Acc &$\Delta$Acc &FP &FN \\
        \midrule
    TuSimple &92.52 &-3.52\% &0.0717 &0.0883 \\
    \midrule
    Metric &F1 &$\Delta$F1 &Recall &Precision \\
    \midrule
    CULane &73.80 &-2.56\% &72.41 &75.26  \\
    \midrule
    SDLane &59.99 &-32.98\% &56.27 &64.25  \\
    \midrule

    \end{tabular}
    }
\end{minipage}
\begin{minipage}[h]{0.48\textwidth}
\makeatletter\def\@captype{table}
\caption{Comparison on \textit{ELM} trained by the EIE loss and the MSE loss.}
\label{tab.losscompare}
    \centering
    \resizebox{0.9\textwidth}{!}{
    \begin{tabular}{@{}c|c|ccc@{}}
    \toprule
    Metric &Loss &Accuracy &FP &FN \\
    \midrule
    \multirow{2}{*}{TuSimple} &MSE &95.18 &0.0501 &0.0624 \\
     &EIE &\bf{95.92} &\bf{0.0295} &\bf{0.0444} \\
    \midrule
    Metric &Loss &F1 &Recall &Precision \\
    \midrule
    \multirow{2}{*}{CULane} &MSE &72.88 &71.17 &74.68 \\
     &EIE &\bf{75.74} &\bf{72.48} &\bf{79.32} \\
    \midrule
    \multirow{2}{*}{SDLane} &MSE &79.49 &79.59 &79.38 \\
     &EIE &\bf{86.34} &\bf{86.92} &\bf{85.77} \\
    \midrule
    \end{tabular}
    }
\end{minipage}

\section{Conclusion}


We propose an efficient geometry-flexible lane detection ElasticLaneNet, which excels at detecting lanes with complex geometric structures, with effectiveness and efficiency simultaneously. In our framework, a novel implicit representation of lanes is designed, \textit{i.e.} zero-contour of \textit{ELM}, which enables superior lane geometries learning on cross, large curves, Y-shaped lanes, etc. We apply the EIE loss to inform the \textit{ELM} learning, which incorporates the long range information of lanes in the whole image space as well as retains attention on local features. CRSN are jointly trained with \textit{ELM}, and TB as well as AFR modules are constructed to improve the feature fusion ability. In the experiments, our ElasticLaneNet outperforms the current models on structurally diverse data SDLane. Ablation studies show the functions of different modules in our network, and the effectiveness of EIE-informed implicit training strategy in challenging cases. In future work, more careful design in network can be developed to combine the EIE loss and the implicit/explicit lanes representation to achieve further improvement.

\bibliographystyle{unsrt}  
\bibliography{references} 

\clearpage
\setcounter{page}{1}

\section*{Appendix}
This Appendix presents supplementary contents of the main paper. In Sec.~\ref{sec:networkstructure}, more details of the network structure of our proposed ElasticLaneNet are illustrated. In Sec.~\ref{sec:mimplementation}, more details in the implementation of our ElasticLaneNet are provided. Section~\ref{sec:explicitElasticlanenet} provides further information of the explicit implementation of ElasticLaneNet$^{pw}$. Section~\ref{sec:failureCU} gives some discussion on the results different from ground truth on CULane. In Sec.~\ref{sec:comparison}, the settings of other models in the comparison experiments on SDLane are presented.


\setcounter{section}{0}
\renewcommand\thesection{\Alph{section}}
\renewcommand \thesubsubsection {\thesubsection \arabic{subsection}}

\section{More Details of the Network Structure}\label{sec:networkstructure}
This section provides more details of the network structure of our ElasticLaneNet shown in Fig.~\ref{fig:networkarchitecture} in Sec.~\ref{sec:methodology} in the main text. In Fig.~\ref{fig:networkarchitecture}, the backbone consists of an Encoder ResNet34 and a Feature Pyramid Network (FPN),
in which the down-sampling processes are indicated by the downward red arrows, the up-sampling processes by the green arrows, and the features concatenation processes by the gray arrows; the main stream is from the backbone to the \textit{ELMM}, while the auxiliary steps include the Transformer Bottleneck layer (TB), classification sub-network (CSN), the range sub-network (RSN) and the auxiliary feature refinement (AFR). These auxiliary processes are framed (or connected) by dotted lines (or arrows), which means they can be removed.

After applying CSN and RSN, in addition to using the loss functions $\mathcal{L}_{exist}$ (lane existence loss) and $\mathcal{L}_{range}$ (the binary-cross entropy loss), the outputs are taken element-wise multiplication with the outputs from \textit{ELMM}, i.e. Pred$_1$, to jointly train and learn the implicit \textit{ELM}s $\Psi_{p_1}^k$'s, $k=1,2,...,N$, with the size of $N\times M\times w$. In AFR, the auxiliary loss $\mathcal{L}_{aux}$  is applied on Pred$_i$, namely $\Psi_{p_i}^k$, $i=2,3$. The feature fusion (FF) in AFR module means adopting convolution on the concatenation of $F_2$ and $F_3$ with $P_1$, the last layer from FPN, to obtain Pred$_1$. $F_2$ and $F_3$ here are the features up-sampled from $P_2$ and $P_3$, having the same size as $P_1$, i.e., $40\times 100$.


\section{More Details of Implementation}\label{sec:mimplementation}
\subsection{Parameters Setting}
During our training, the batch size is set to be 24 to 32, the optimizer is AdamW~\cite{loshchilov2018fixing}, the learning rate is about $3\times 10^{-4}$ for both three datasets. The data augmentation includes resize, flip, channel shuffle, random brightness or saturation change, affine transformation, etc., same as the setting in ~\cite{zheng2022clrnet}.

\subsection{Data Preparation
and Model Evaluation}
In the experiment of SDLane, we first reorganize the data structure into CULane format~\cite{pan2018spatial}, and use the same official evaluation metrics as those in CULane to evaluate the performance. The evaluation code of these two datasets is the same as that in~\cite{tabelini2021keep}, which is a python re-implementation of the official C++ code.

\setcounter{figure}{0}
\renewcommand{\thefigure}{\alph{figure}}
\begin{figure*}[ht]
  \centering
   \includegraphics[width=0.9\textwidth]{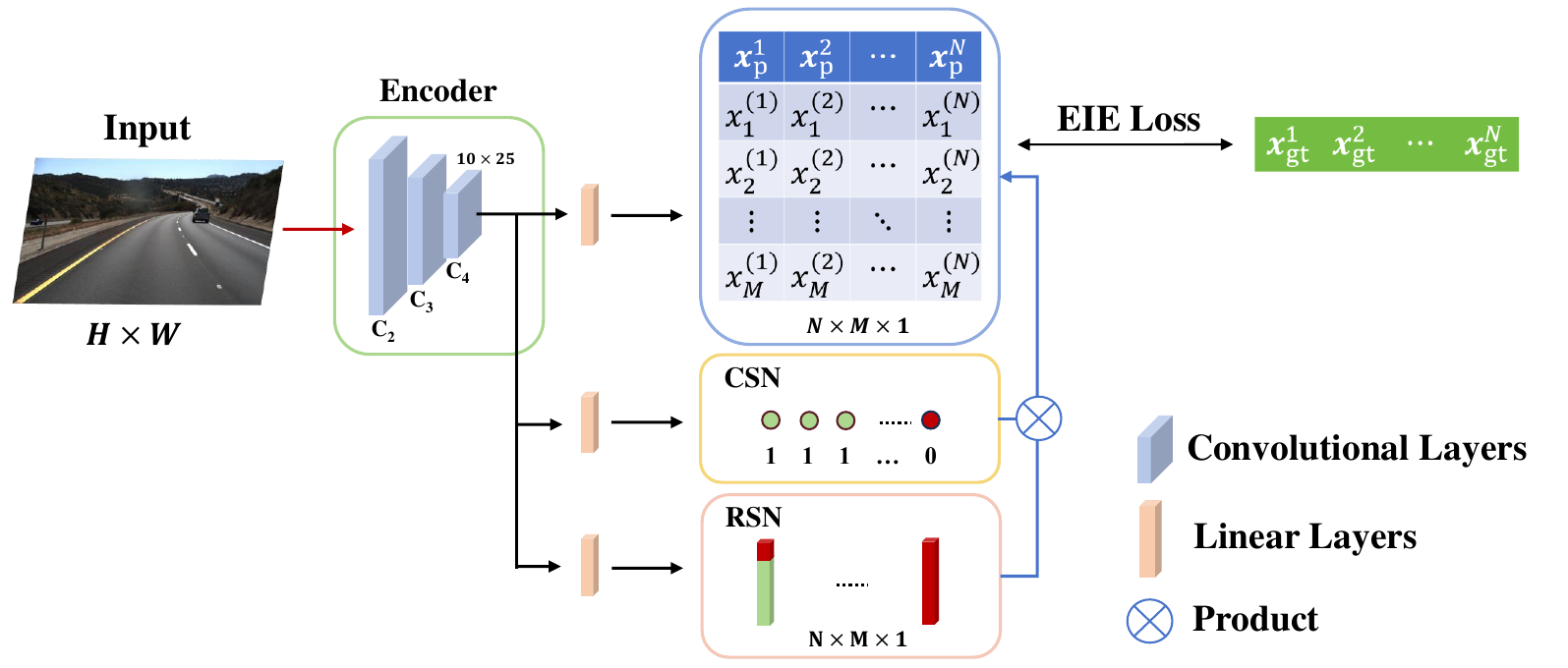}
   \caption{Network architecture of ElasticLaneNet$^{pw}$.}
   \label{fig:elasticlanenetpw}
\end{figure*}

\section{More Details of the Explicit ElasticLaneNet$^{pw}$}\label{sec:explicitElasticlanenet}
In this section, we describe more details of ElasticLaneNet$^{pw}$, which is an alternative EIE loss based implementation of ElasticLaneNet. Compared to the implicit implementation of ElasticLaneNet presented in the main text, from which we obtains the lane points from the zero contours of the output \textit{ELM}s, the ElasticLaneNet$^{pw}$ predicts the $x$-coordinates of N lanes $\left(\boldsymbol{\gamma}_p^{1}, \dots, \boldsymbol{\gamma}_p^{N}\right)$ directly from the network, \textit{i.e.} $\left(\boldsymbol{x}_p^{1}, \dots, \boldsymbol{x}_p^{N}\right)$, where $ \boldsymbol{x}_p^{k} = \left(x_{1}^{\left(k\right)},x_{2}^{\left(k\right)},\ldots,x_{M}^{\left(k\right)}\right)^\top$; see Fig.~\ref{fig:elasticlanenetpw}. The ground truth lanes are $ \boldsymbol{\gamma}_{gt}^{k} $, and coordinates on them are $ \left(\boldsymbol{x}_{gt}^{k}, \boldsymbol{y}_{gt}^{k}\right), k=1,\dots,N$. 

As mentioned in Sec.~\ref{subsec:EIE}, the elastic interaction energy (EIE) of two groups of pair-wise open curves is 
\begin{equation}\label{eq.eie}
E =\frac{1}{8 \pi} \int_{\gamma} \int_{\gamma^{\prime}} \frac{d \boldsymbol{l} \cdot d \boldsymbol{l}^{\prime}}{r},
\end{equation}
where vector $d \boldsymbol{l}$ represents line element on curves $\gamma$ with tangent direction $\boldsymbol{\tau}$, \textit{i.e.} $d\boldsymbol{l}=\boldsymbol{\tau} d l$, $d l$ is the arc length, $\gamma^{\prime}$ denotes the curves with another parameter, \textit{i.e.} $\gamma = \gamma \left( s \right)$ and $\gamma^{\prime} = \gamma \left( s^{\prime} \right)$, $s$ ($s^\prime$) is the parameter. The coefficient $\frac{1}{8\pi}$ is set before the EIE so that the Fourier transform of EIE has no coefficient:  $\mathcal{L}_{eie}=\sum\limits_{m,n} \sqrt{m^2+n^2}\left|d_{mn}\right|^2$. 

In order to apply an explicit method, the energy can be rewritten using the relationship of $d \boldsymbol{l}=\boldsymbol{\tau} \delta(\gamma) \mathrm{d}x\mathrm{~d}y$, as 
\begin{equation}\label{eq.eie3}
E\left(x,y\right)=\frac{1}{8 \pi} \int_{\mathbf{R}^2} \delta(\gamma) \mathrm{d} x \mathrm{~d} y \int_{\mathbf{R}^2} \frac{\boldsymbol{\tau} \cdot \boldsymbol{\tau}^{\prime}}{r} \delta\left(\gamma^{\prime}\right) \mathrm{d} x^{\prime} \mathrm{d} y^{\prime},
\end{equation}
where $\delta(\cdot)$ is a Delta function.
In the implementation, a regularized delta function is required to smear out the singularities.

The velocity field of $\gamma$ as well as the negative gradient descent direction of the EIE is:
\begin{equation}\label{eq.veie}
\begin{aligned}
\gamma_t(x, y)&=-\frac{\delta E}{\delta \gamma} \\
& =\left(-\frac{1}{4 \pi} \int_\gamma \frac{\boldsymbol{r} \times d \boldsymbol{l}}{r^3}\right) \times \tau \\
&=\left(\frac{1}{4 \pi} \int_\gamma \frac{\boldsymbol{r} \cdot \boldsymbol{n}_\gamma}{r^3} d l\right) \boldsymbol{n} \\
&= \left(\frac{1}{4 \pi} \int_{\mathbf{R}^2} \frac{\boldsymbol{r} \cdot \boldsymbol{n}_{\gamma^{\prime}}}{r^3} \delta(\gamma^{\prime} ) \mathrm{d}x^{\prime}\mathrm{d}y^{\prime} \right) \boldsymbol{n},
\end{aligned}
\end{equation}
where $\boldsymbol{\tau}_\gamma$ and $\boldsymbol{n}_\gamma$ are the unit tangent and normal vector of parameterized curve $(x(s),y(s))$, and $\boldsymbol{\tau}$ and $\boldsymbol{n}$ are the unit tangent and normal vector of curve on $(x,y)$ over the whole image domain. In the derivation, $d \boldsymbol{l} = \boldsymbol{\tau}_\gamma dl$, $\hat{z} \times \boldsymbol{\tau} = \boldsymbol{n}$ and $\hat{z} \times \boldsymbol{\tau}_\gamma = \boldsymbol{n}_\gamma$, where $\hat{z}$ is the unit vector in the $+z$ direction. 

In Eq.~\ref{eq.veie}, the $\gamma$ consists of the pair-wise ground truth $\gamma_{gt}$ and the prediction lane $\gamma_{p}$. The prediction lane $\gamma_{p}$ has opposite orientation to the ground truth $\gamma_{gt}$ as mentioned in Sec.~\ref{subsec:EIE}. The velocity of the prediction is:
\begin{equation}\label{eq.veie.gamma}
{\gamma_p}_t(x, y)= -\left(\frac{1}{4 \pi} \int_{\mathbf{R}^2} \frac{\boldsymbol{r} }{r^3} \cdot (\boldsymbol{n}_{\gamma_{gt}^{\prime}}\delta(\gamma_{gt}^{\prime} ) - \alpha \boldsymbol{n}_{\gamma_{p}^{\prime}}\delta(\gamma_{p}^{\prime} )) \mathrm{d} x^{\prime} \mathrm{d} y^{\prime} \right) \boldsymbol{n}_{\gamma_{p}},
\end{equation}
where $\alpha$ is the same hyper-parameter as mentioned in Eq.~\ref{eq.leie}.

In order to simplify the calculation, the regularized delta function chosen here is $\delta_\sigma\left(x-x_m\right)=\left\{\begin{array}{ll}\frac{1}{2 \sigma}, & \text { if } x \in \mathrm{U}\left(x_m, \sigma\right) \\ 0, & \text { if } x \notin \mathrm{U}\left(x_m, \sigma\right)\end{array}, m=1, \ldots, M\right.$, where $x_m$ is one of the $x$-coordinates on a lane. Therefore, the Delta function applied here can be the derivative of a smooth Heaviside function $H_\sigma(\cdot) - 0.5$ mentioned in Sec.~\ref{subsec:ELM}, satisfying equation $\nabla (H_\sigma(\phi)-0.5) = -\delta_\sigma (\gamma)\boldsymbol{n}$. Thus the calculation of EIE loss becomes the same as Eq.~\ref{eq.eie2} and~\ref{eq.leie} in Sec.~\ref{subsec:EIE}.

The velocity field from Eq.~\ref{eq.veie.gamma} becomes:
\begin{equation}
v(x, y) = {\gamma_p}_t(x, y)  = -\frac{1}{4 \pi} \int_{\mathbf{R}^2} \frac{\boldsymbol{r} \cdot \nabla\left(G_t-\alpha \Psi_p  \right)\left(x, y\right)}{r^3} \mathrm{~d} x \mathrm{d} y,
\end{equation}
which can be calculated efficiently via FFT, like Eq.~\ref{eq.veie.fft}.

We guess the degradation of ElasticLaneNet$^{pw}$ on TuSimple and CULane, and the weak performance on geometry-diverse dataset SDLane is because of the non-smooth initial delta functions caused by the randomly initial $x-$sampling. On the contrary, implicit representation of \textit{ELM} can be immediately attracted to the map of ground truth from a more general initial condition. This explicit method may be able to obtain better approximation results by improving the initial sampling distribution or choose a smooth regularized delta function instead.

\begin{figure*}[ht] 
  \centering
  \begin{minipage}[b]{0.96\linewidth} 
  \subfloat[Images]{
    \begin{minipage}[b]{0.32\linewidth} 
      \centering
      \includegraphics[width=\linewidth]{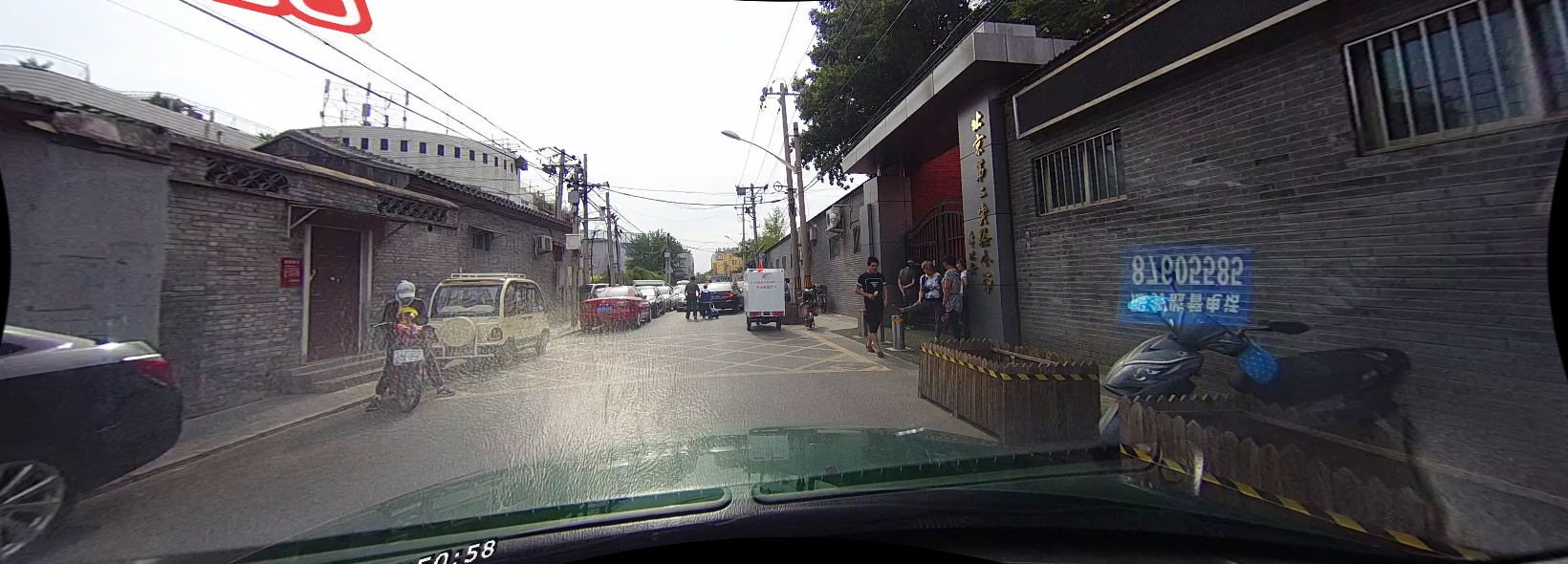}
      \includegraphics[width=\linewidth]{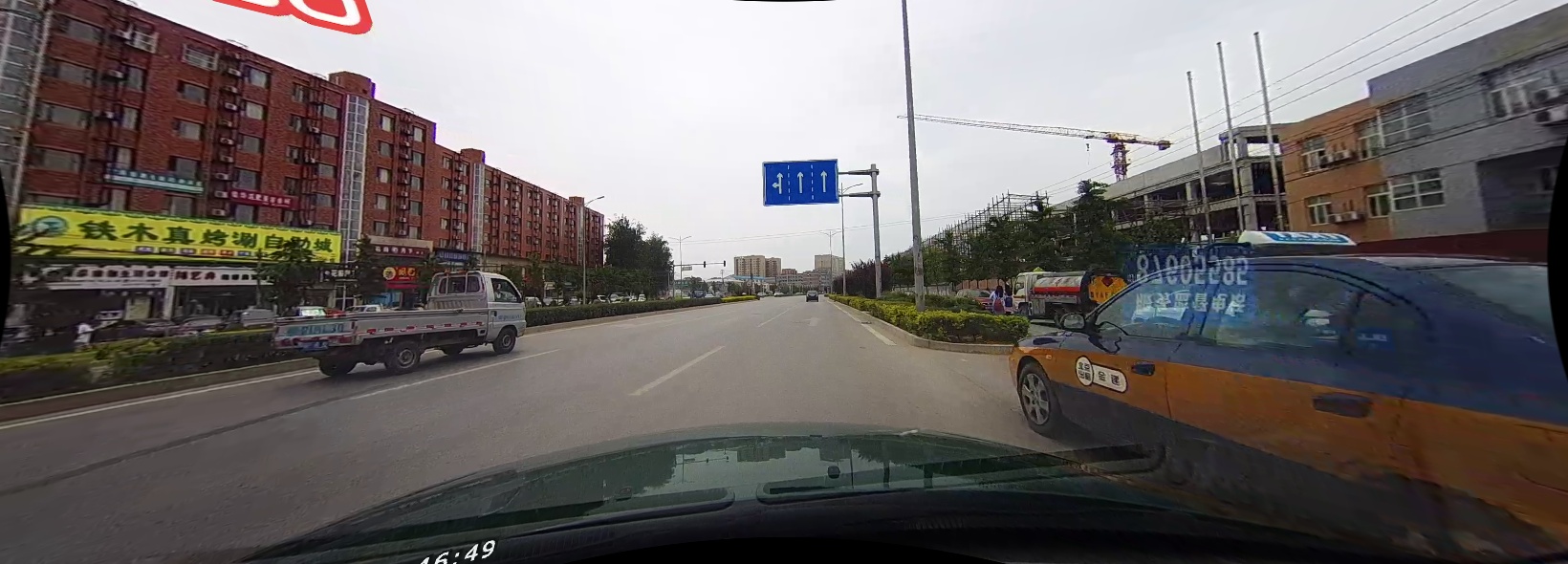}
      \includegraphics[width=\linewidth]{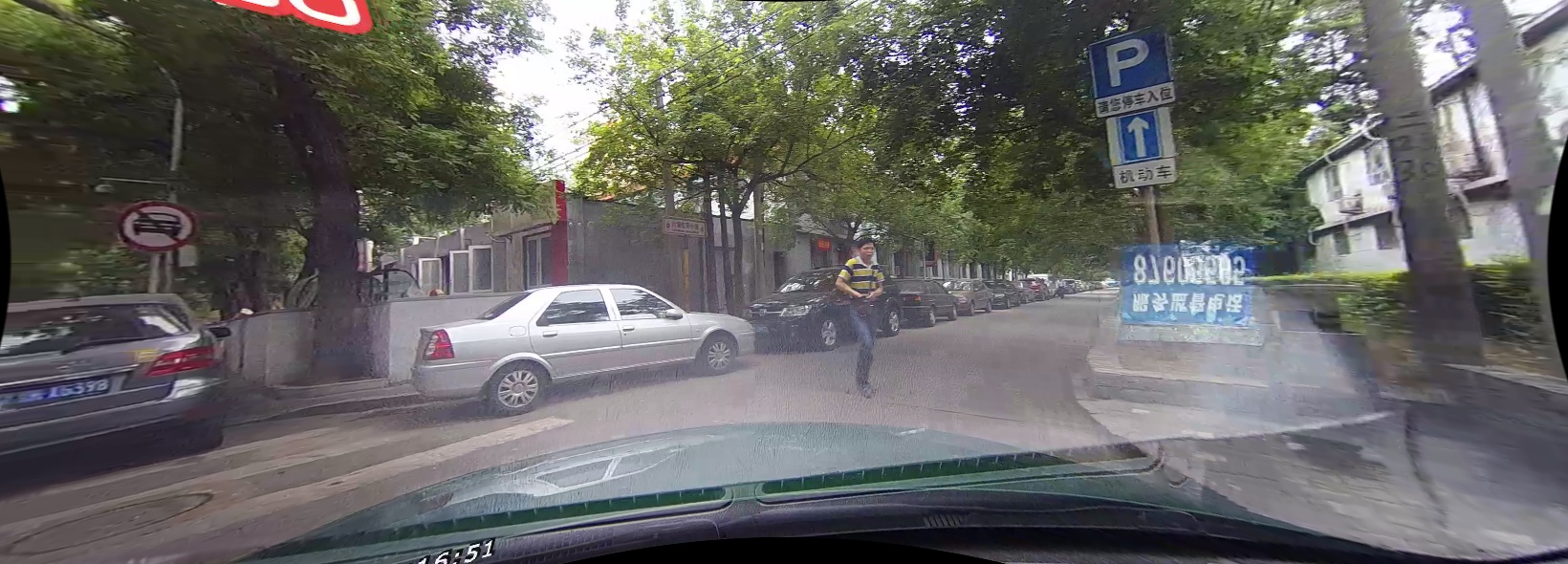}
      \includegraphics[width=\linewidth]{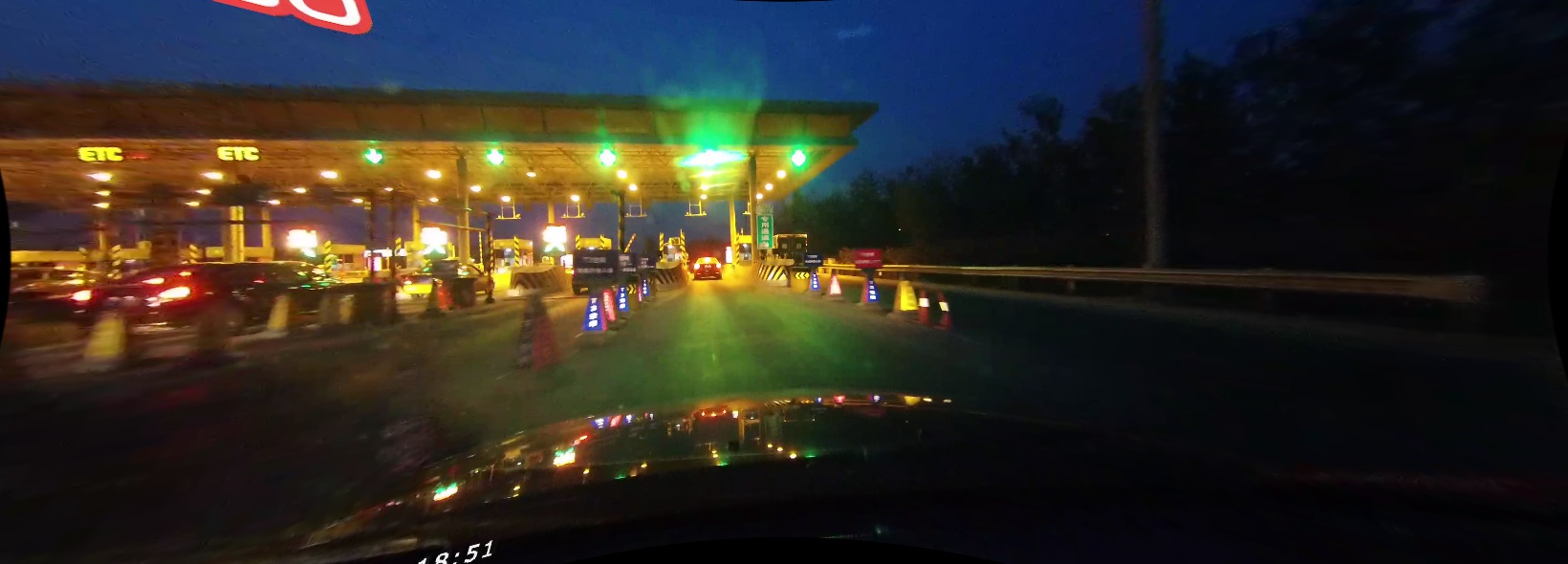}
      \includegraphics[width=\linewidth]{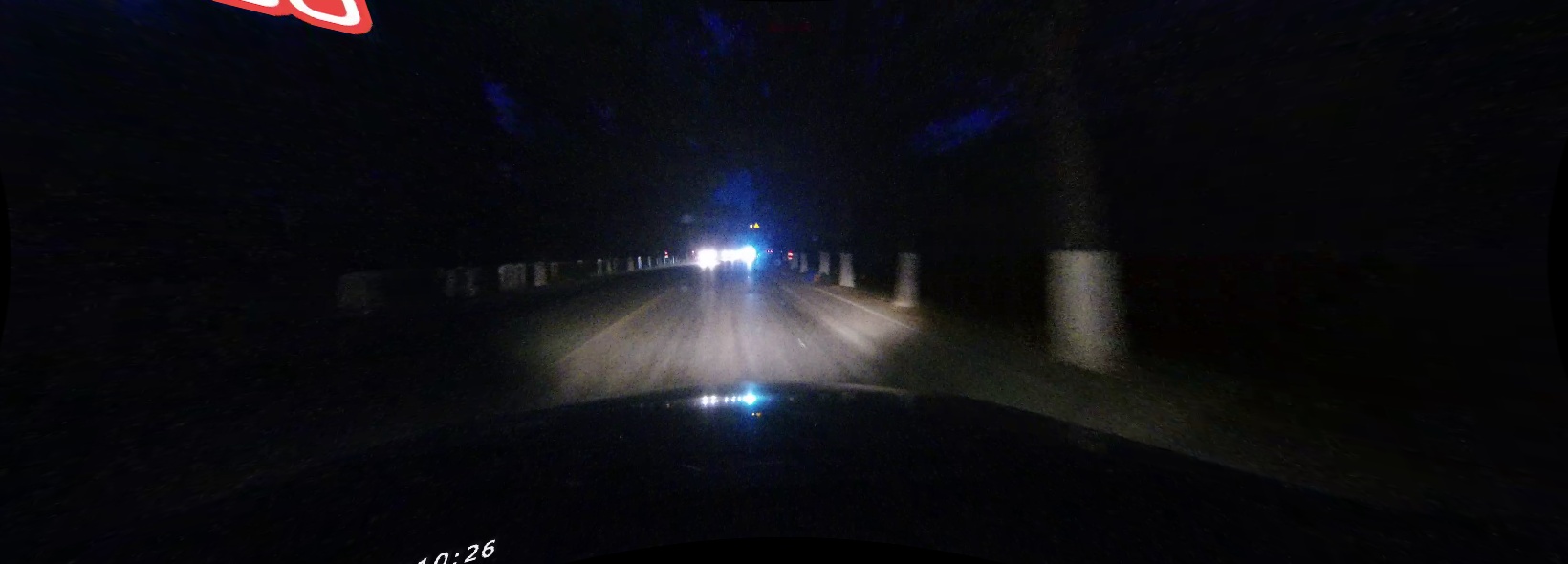}
      \includegraphics[width=\linewidth]{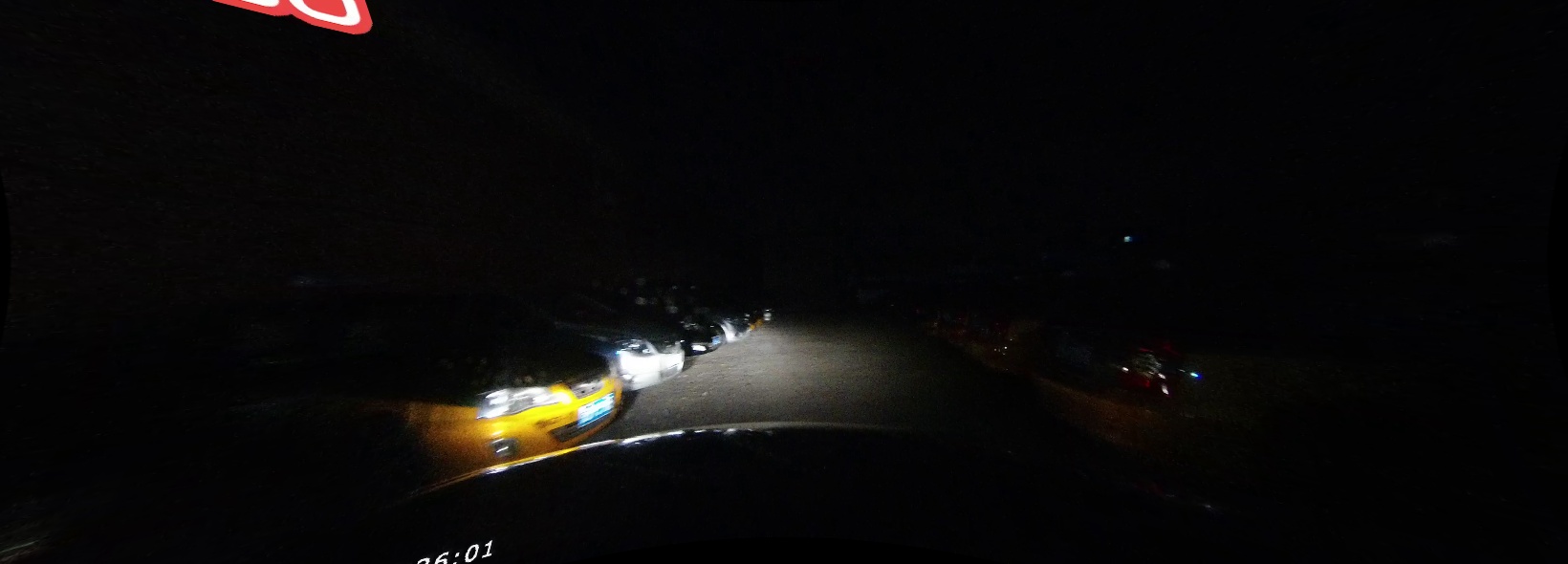}
      \includegraphics[width=\linewidth]{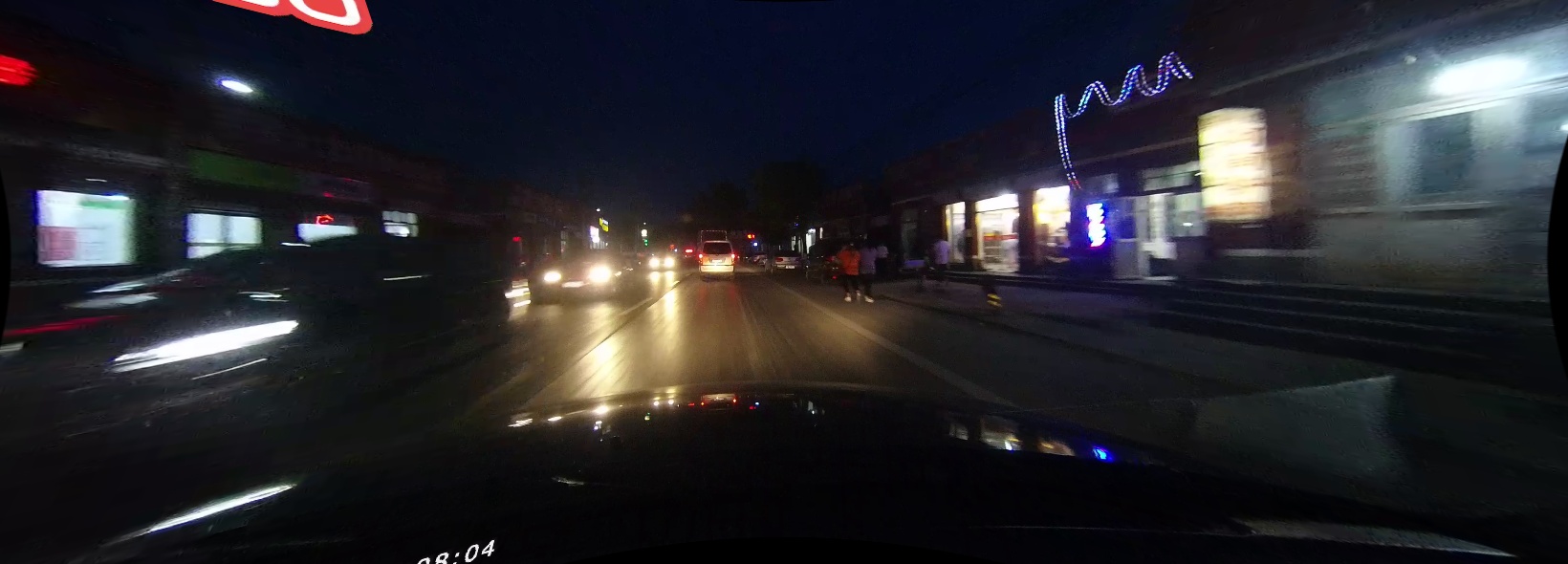}
      \includegraphics[width=\linewidth]{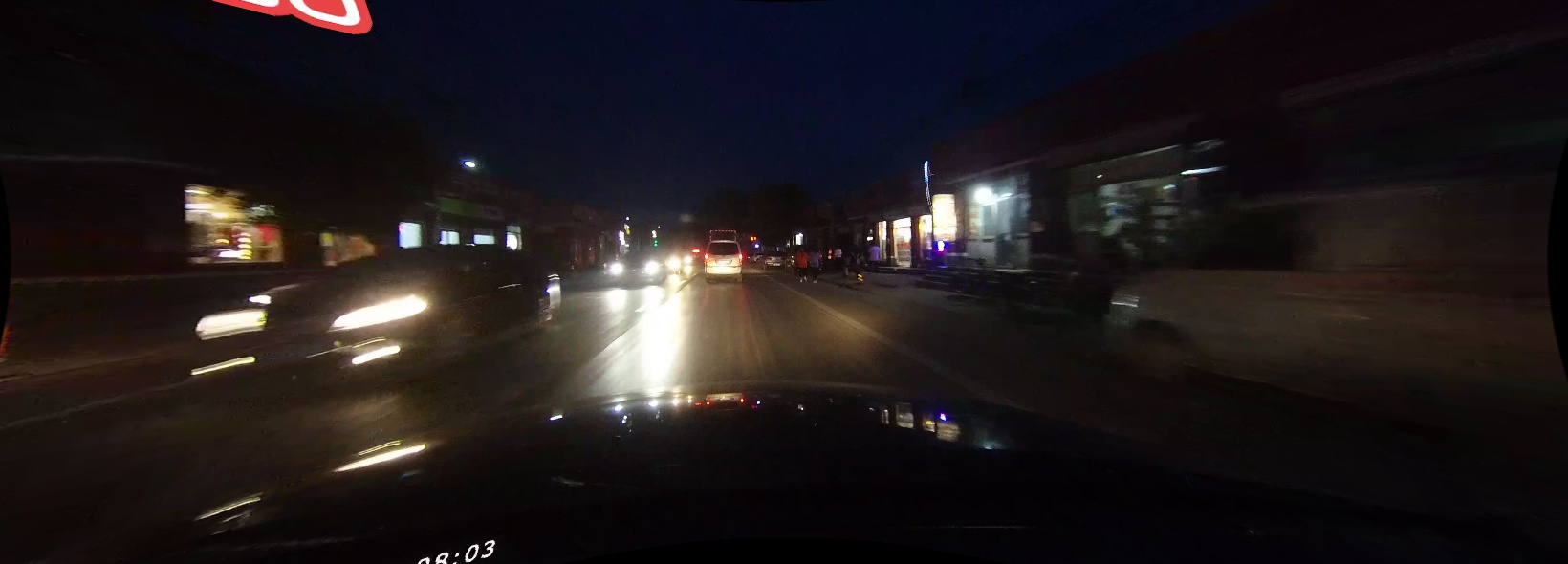}
    \end{minipage}
  }
   \subfloat[Ground Truth]{
    \begin{minipage}[b]{0.32\linewidth}
      \centering
      \includegraphics[width=\linewidth]{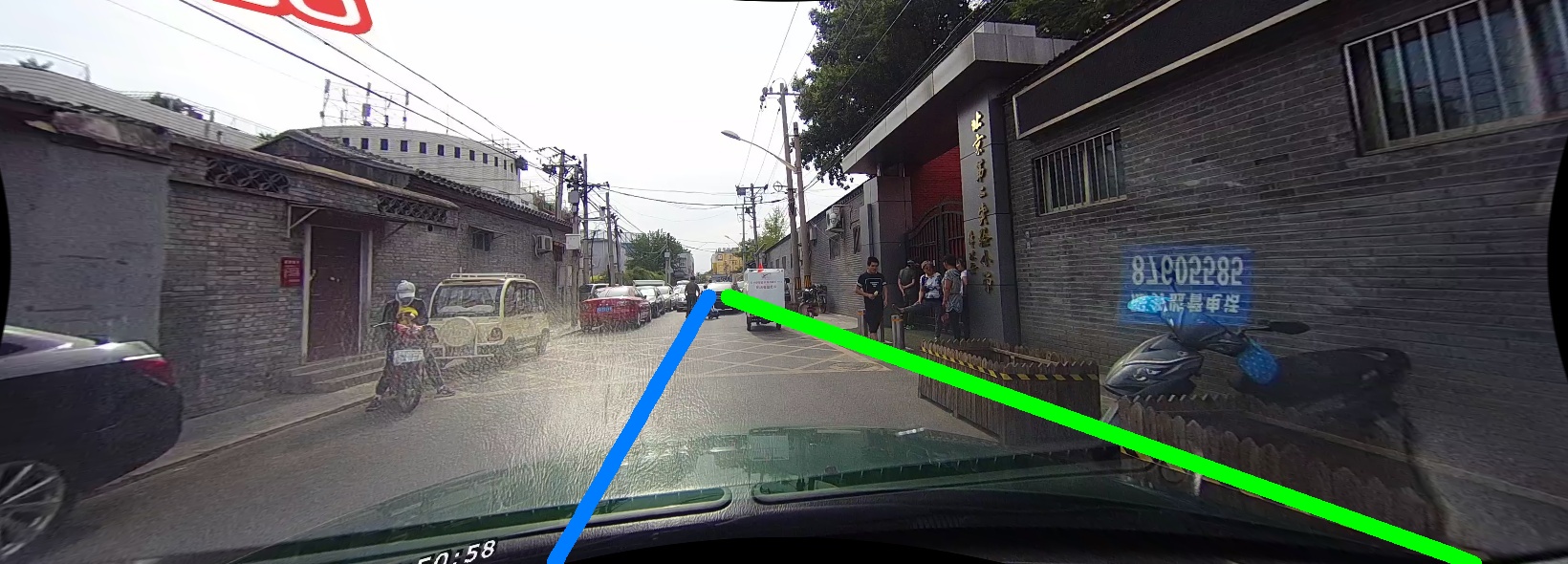}
      \includegraphics[width=\linewidth]{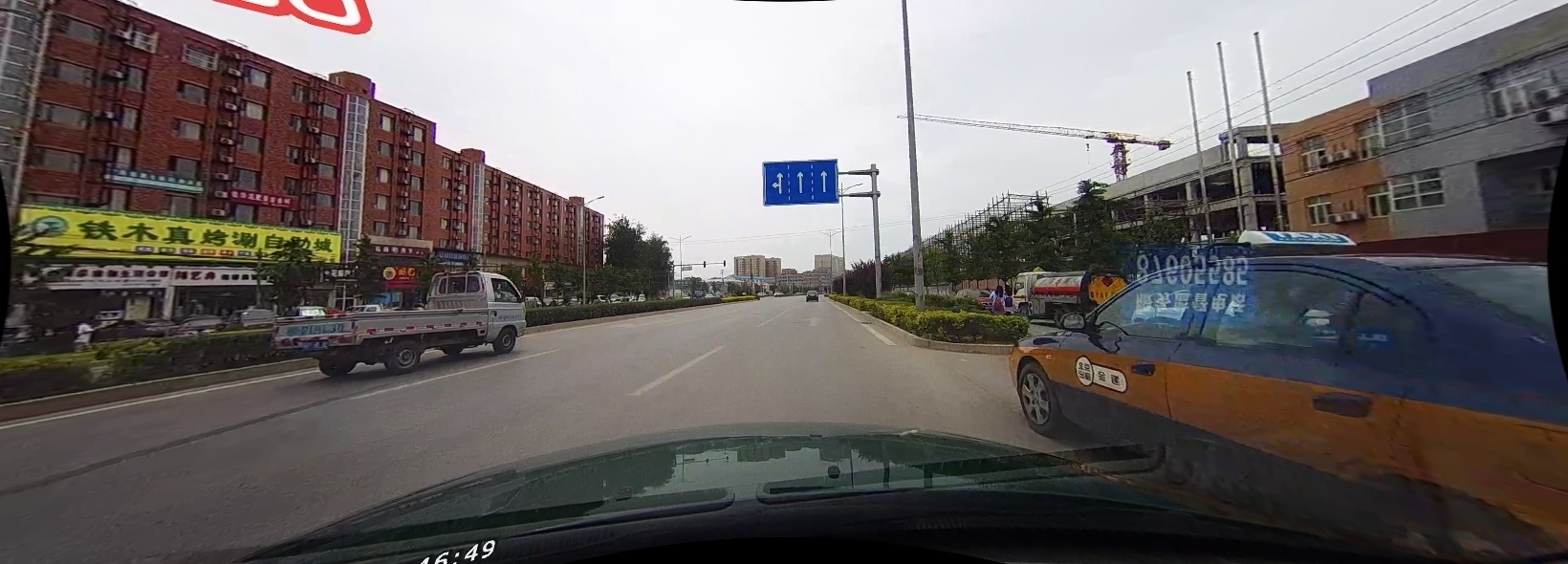}
      \includegraphics[width=\linewidth]{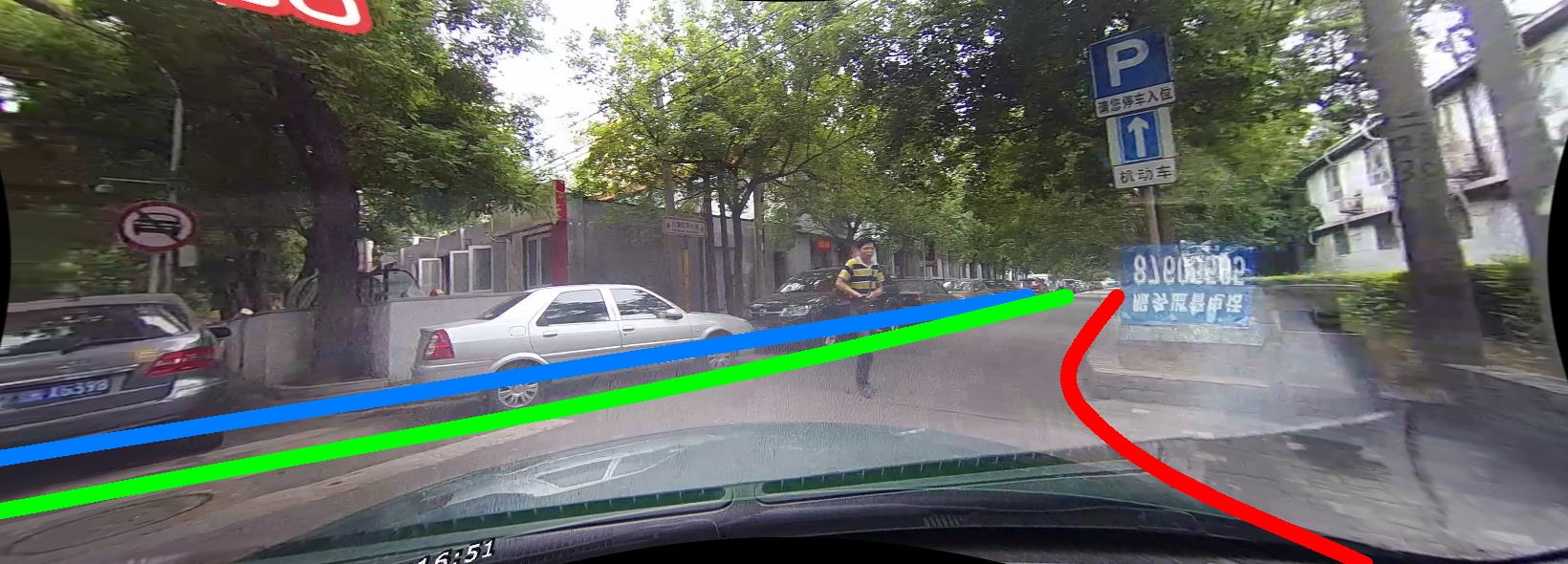}
      \includegraphics[width=\linewidth]
      {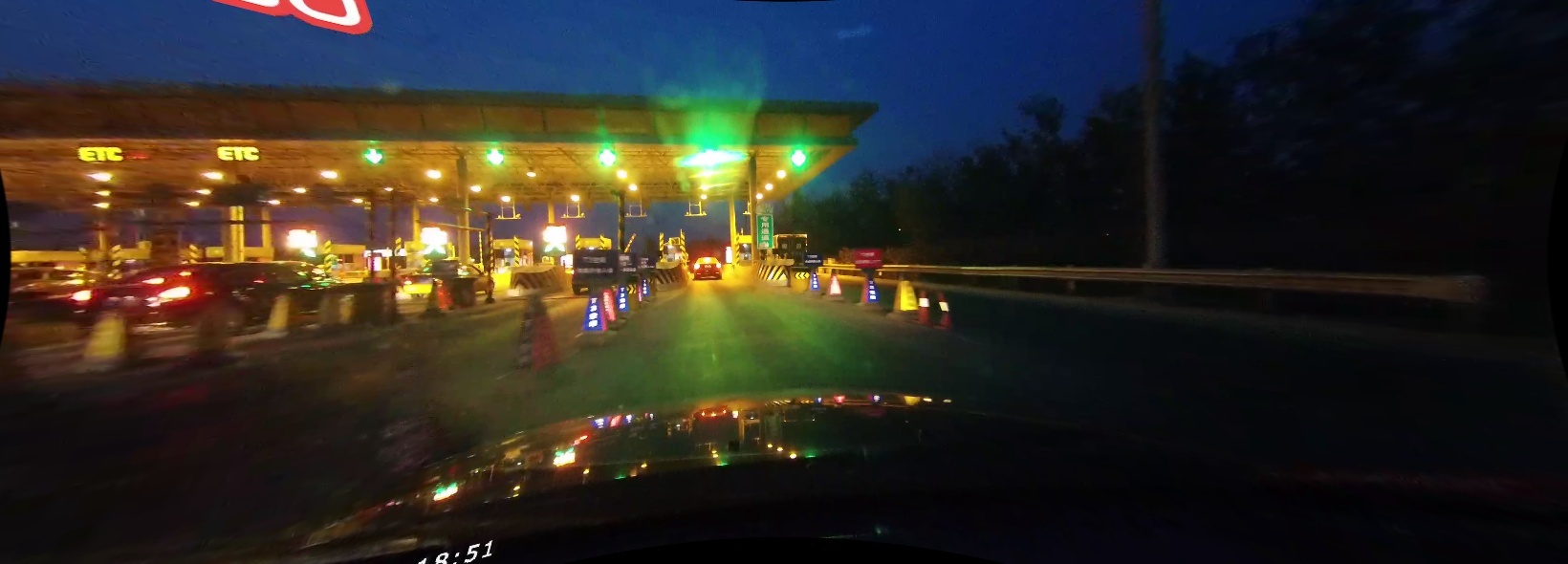}
      \includegraphics[width=\linewidth]{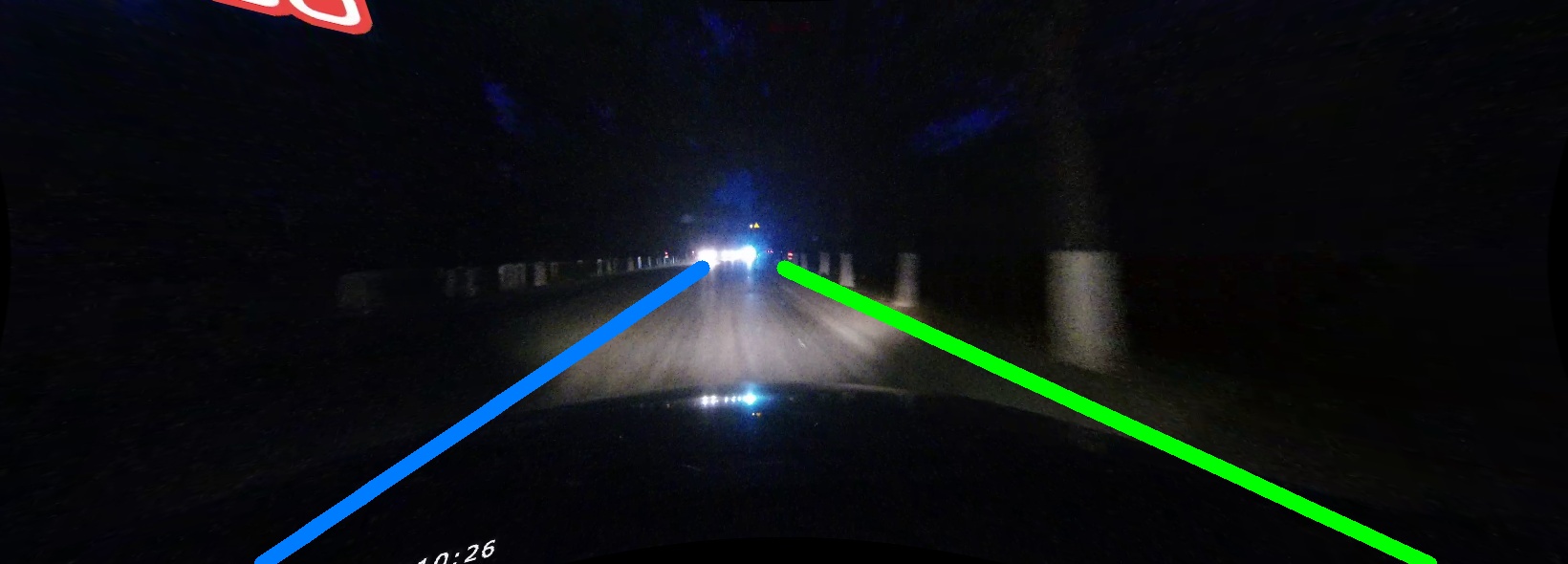}
      \includegraphics[width=\linewidth]{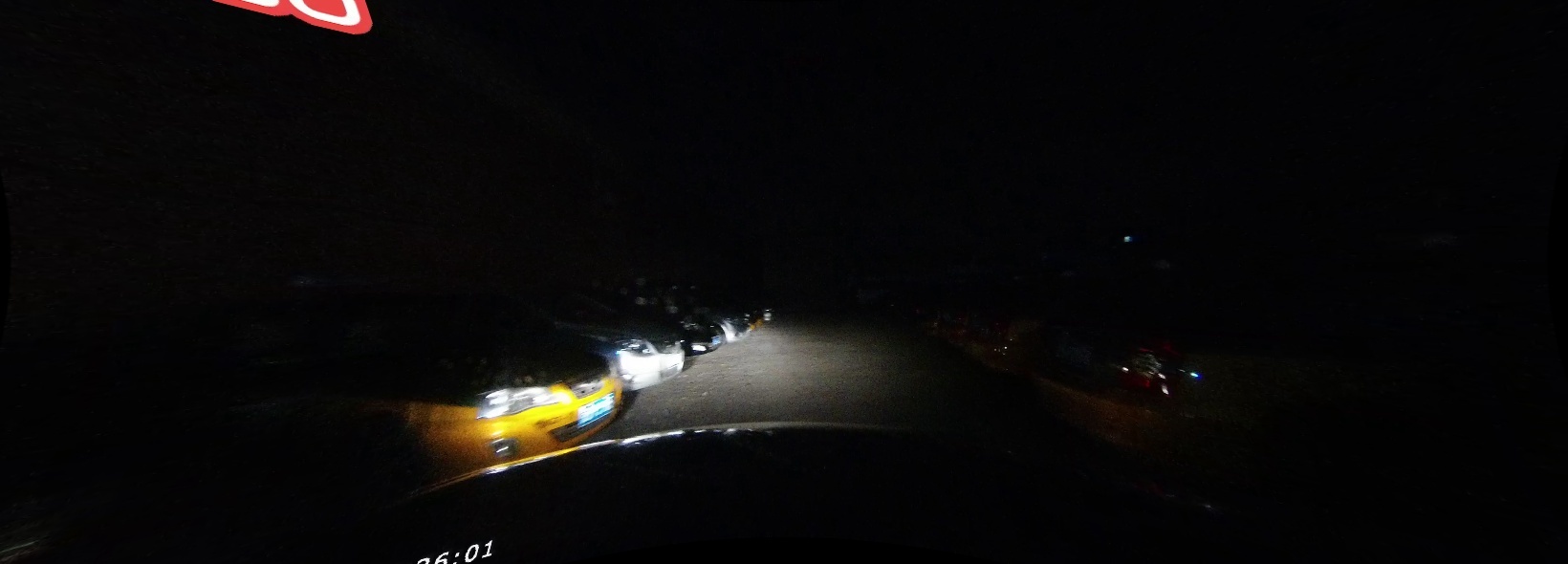}
      \includegraphics[width=\linewidth]{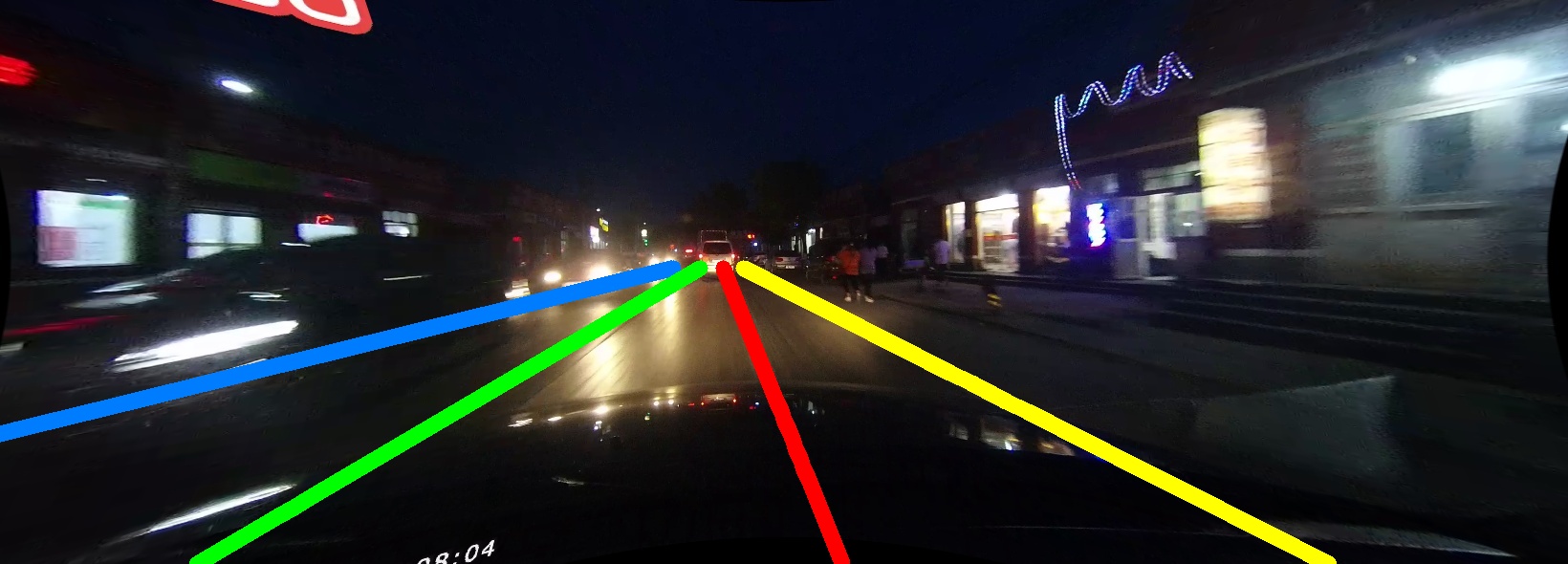}
      \includegraphics[width=\linewidth]{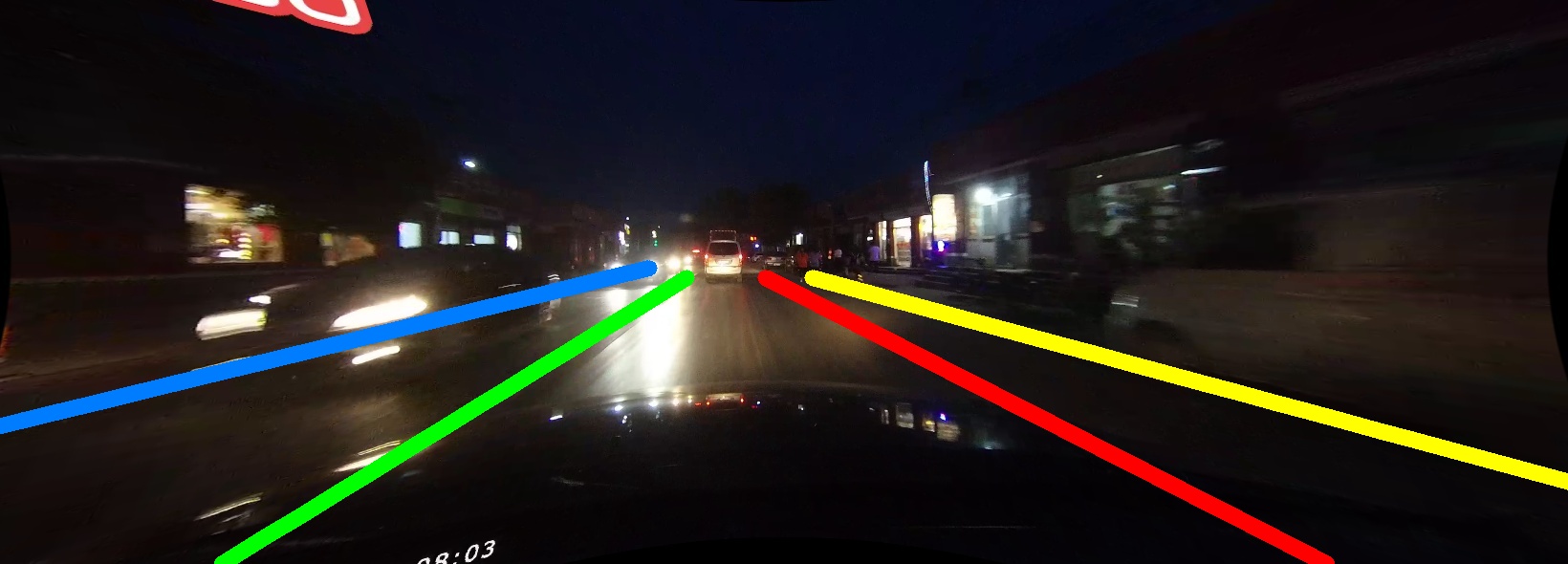}
    \end{minipage}
  }
    \subfloat[Ours]{
    \begin{minipage}[b]{0.32\linewidth}
      \centering
      \includegraphics[width=\linewidth]{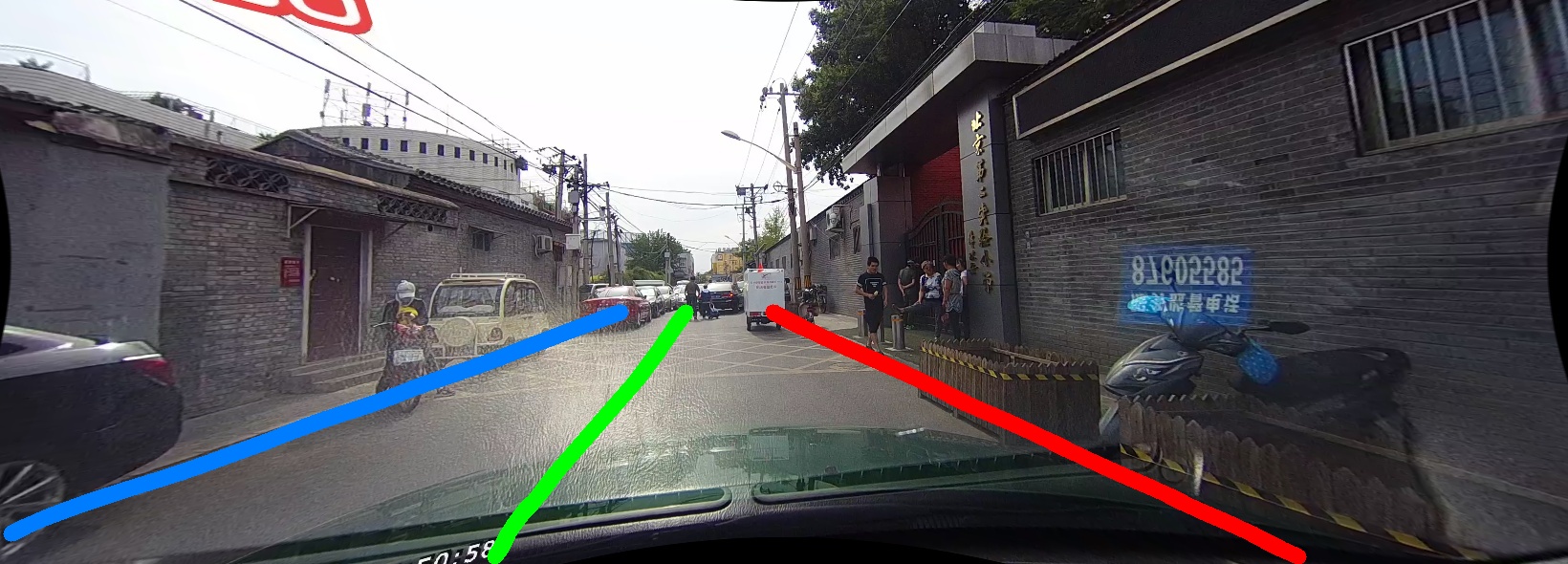}
      \includegraphics[width=\linewidth]{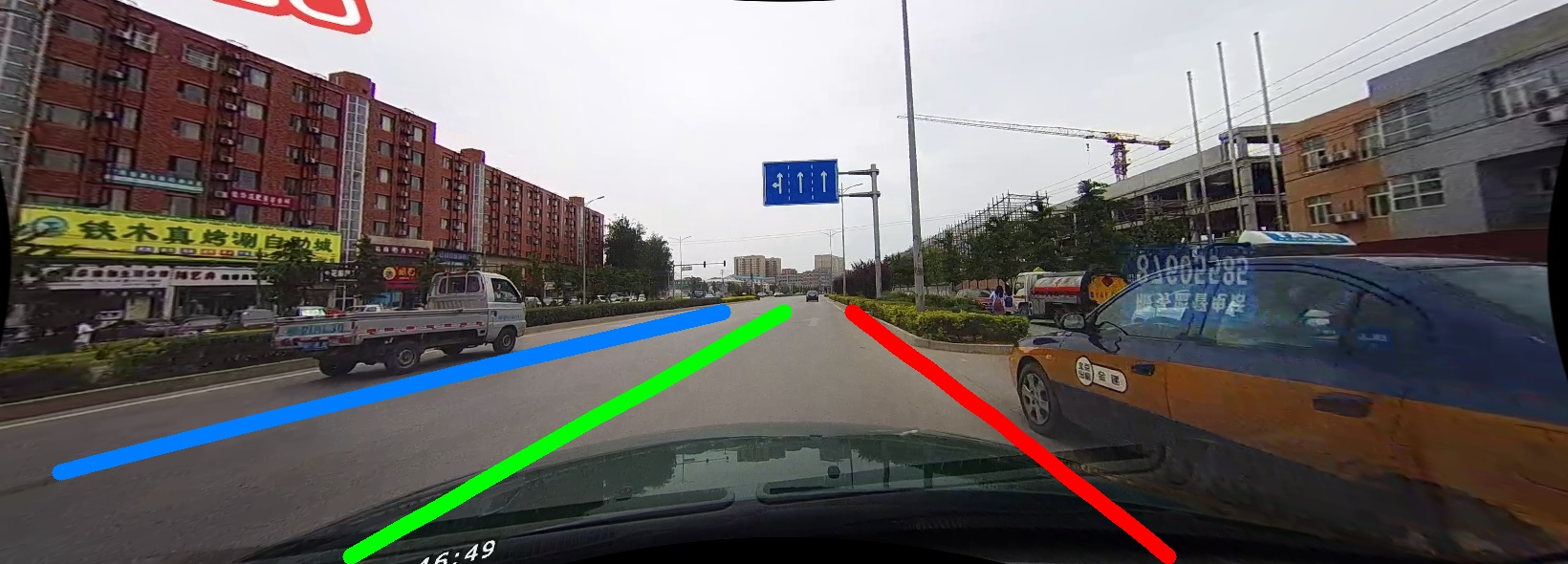}
      \includegraphics[width=\linewidth]{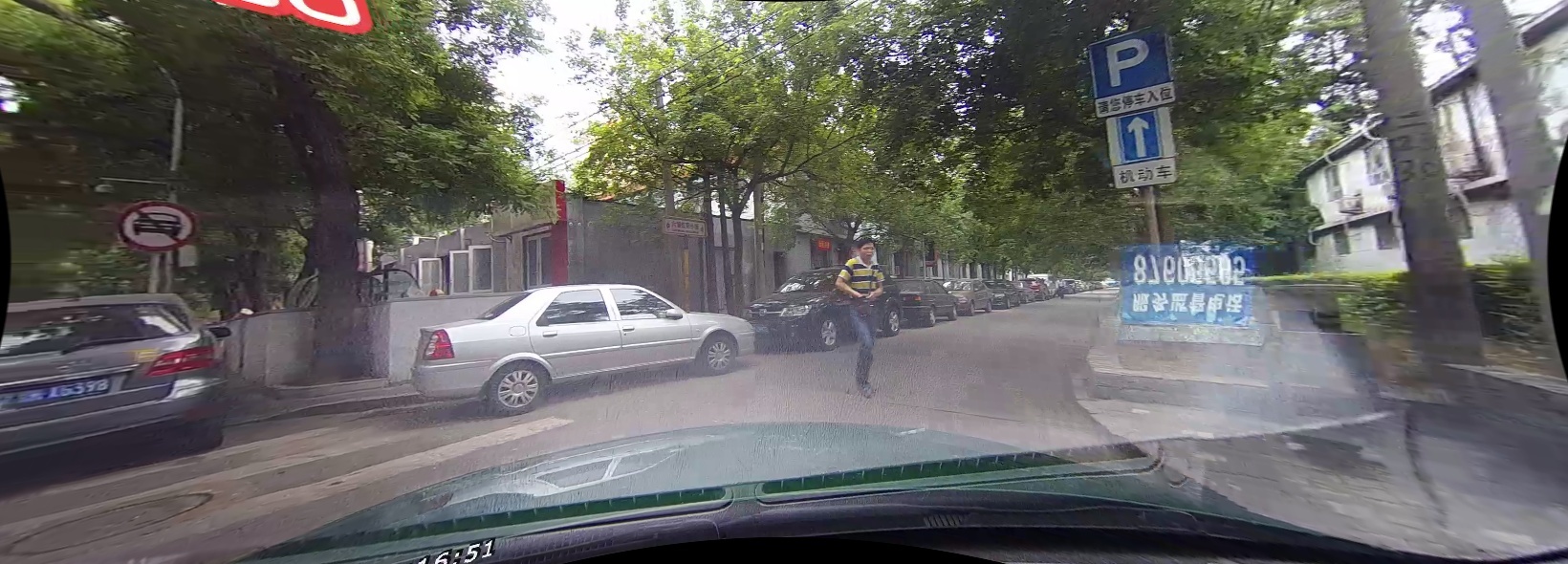}
      \includegraphics[width=\linewidth]{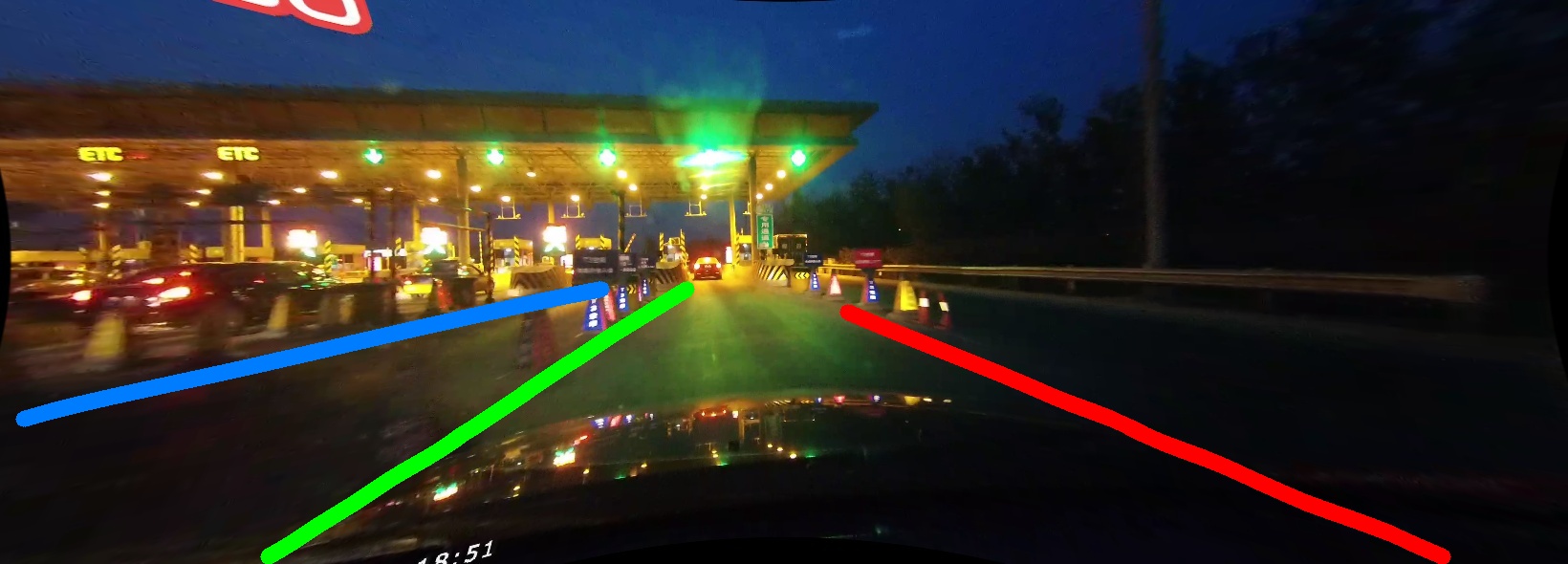}
      \includegraphics[width=\linewidth]{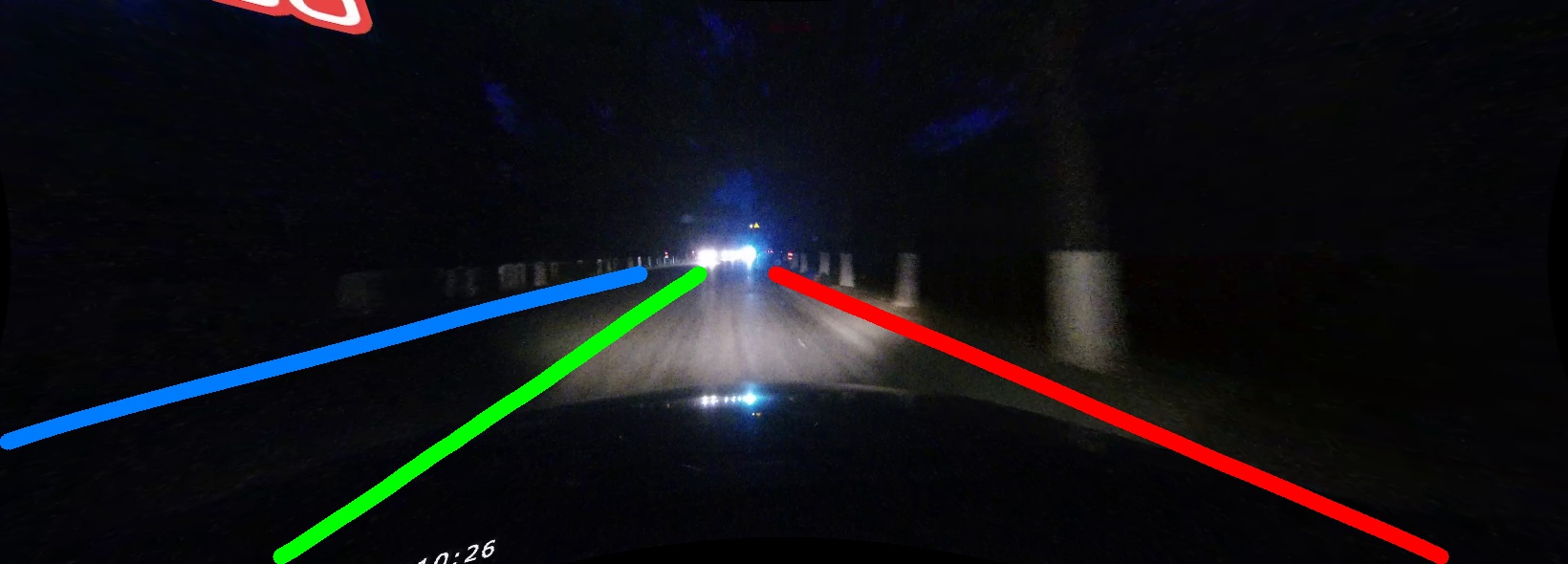}
      \includegraphics[width=\linewidth]{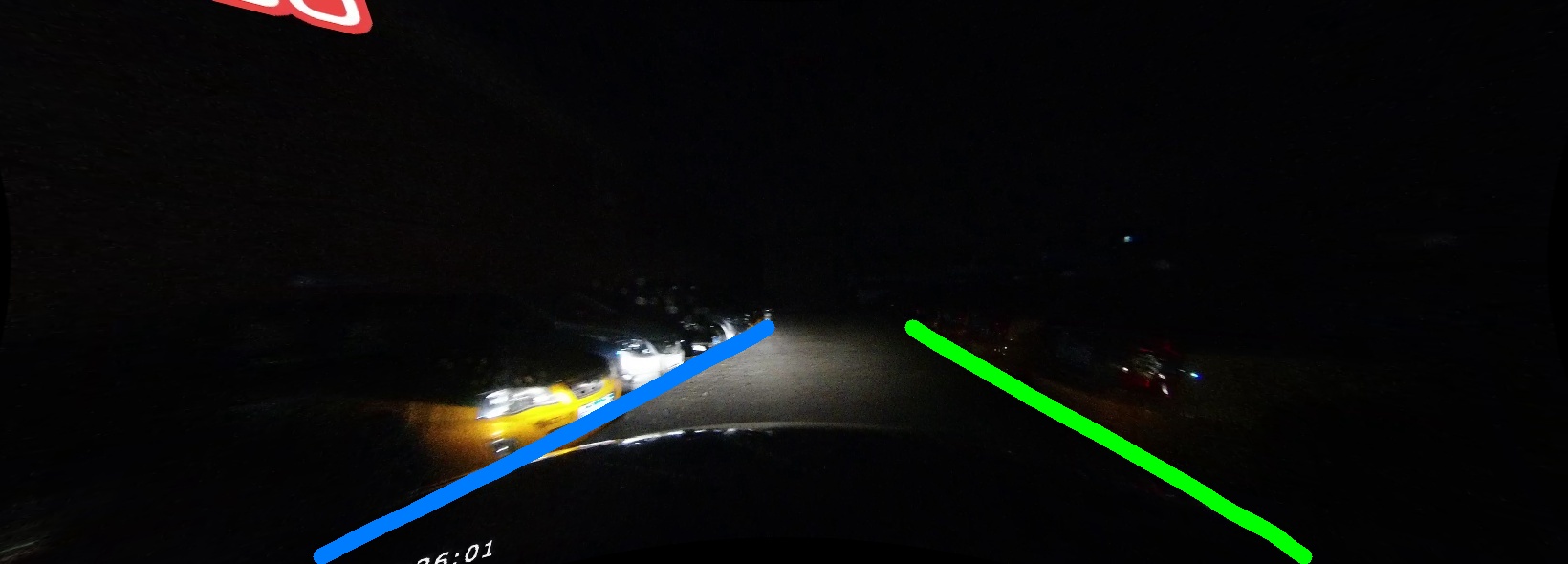}
      \includegraphics[width=\linewidth]{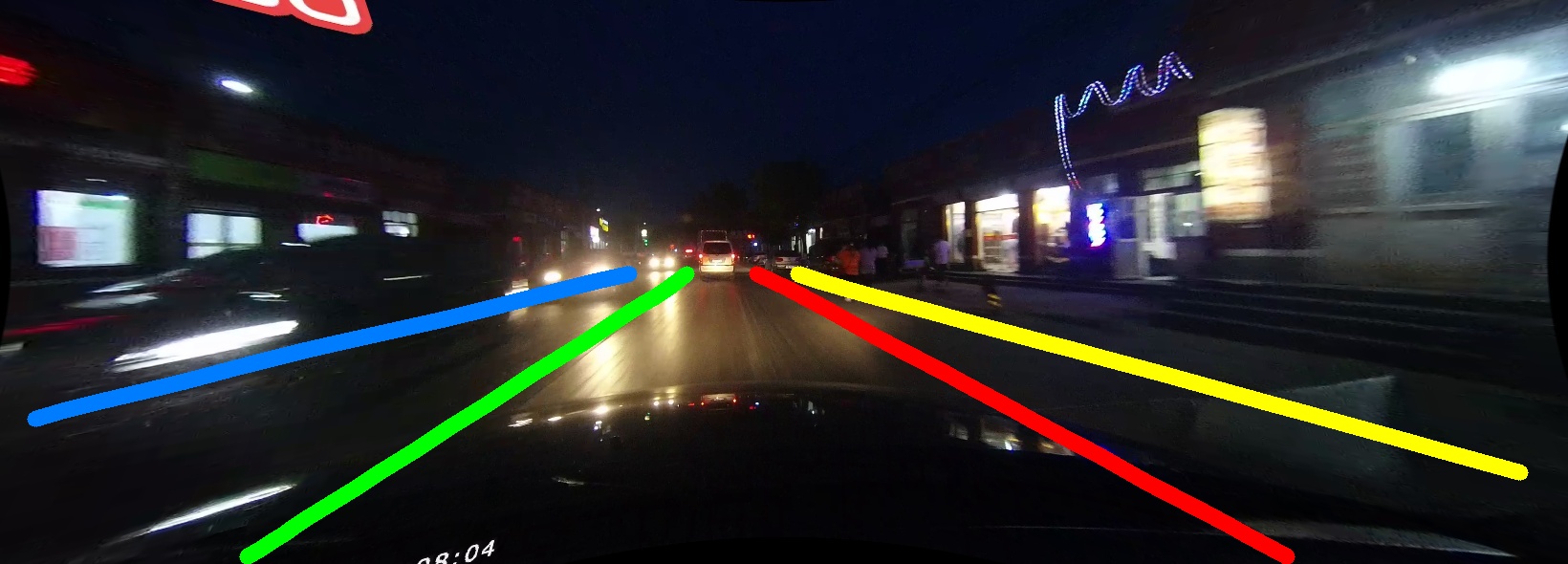}
      \includegraphics[width=\linewidth]{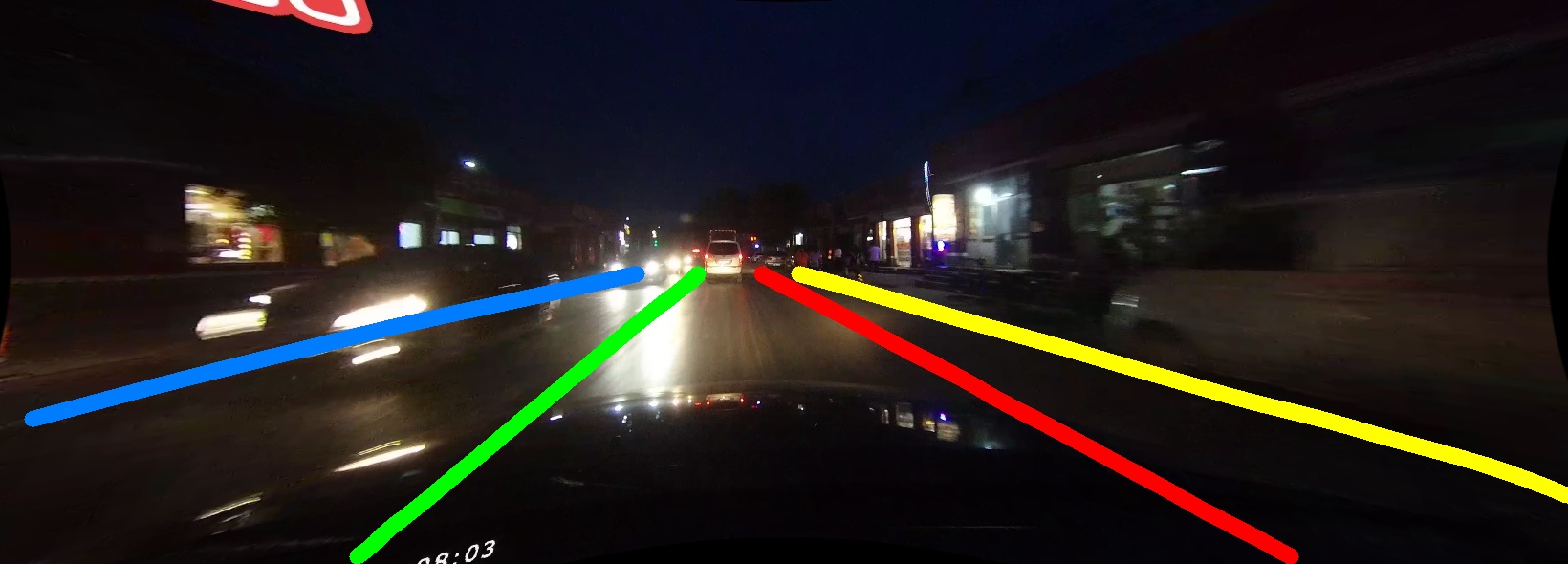}
    \end{minipage}
  }
  \end{minipage}
  \caption{Some results of ElasticLaneNet on CULane dataset that are different from the ground truth.}\label{fig:culanecases}
\end{figure*}

\section{Results Discussion in CULane}\label{sec:failureCU}
In our experiments on CULane, we found that even some of our predictions look acceptable, they are different from the ground truth; see Fig.~\ref{fig:culanecases}. This might because the manual lane marking is sometimes subjective (see the last two rows in Fig.~\ref{fig:culanecases}, which are consecutive frames), and CULane covers a lot of scenes, thus it is difficult to formulate a very uniform labeling standard. Because deep learning methods often learn patterns from large amounts of data, and ElasticLaneNet is sensitive to geometric features in the input images, it can infer more geometric details related to patterns of labeling in the dataset. Therefore, we think both of our test results and the ground truth of CULane displayed in Fig.~\ref{fig:culanecases} are acceptable, although they are different.

This situation also indicates that our approach has the application potential in semi-supervised learning that requires the model has a strong capability on drawing conclusion on labeled data features and producing pseudo labels on unlabeled data, or learn the consistency features from data.

\section{Configuration Settings of Compared Models}\label{sec:comparison}
\begin{figure}[t]
	\centering
        \includegraphics[width=0.6\textwidth]{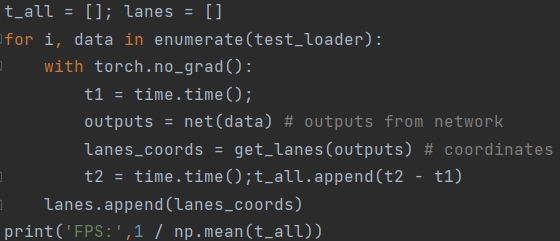}
	\caption{Pseudocode of FPS comparison.}
	\label{fig:fpscode}

\end{figure}

In Sec.~\ref{subsec.compare}, we perform comparison study on highly structure diverse lane data SDLane~\cite{jin2022eigenlanes}. We re-train five of the SOTA methods from the last two years~\cite{abualsaud2021laneaf,qin2022ultra,liu2021condlanenet,feng2022rethinking,zheng2022clrnet} to convergence, and compare the results with those of our ElasticLaneNet; see Tab.~\ref{tab.sd} and Fig.~\ref{fig:sdlaneresults}. The pseudo code of FPS comparison is shown in Fig.~\ref{fig:fpscode}


\subsection{LaneAF.}
LaneAF~\cite{abualsaud2021laneaf} is one of the latest segmentation-based methods with cluster post-processing according to affinity field predictions. Other setting remains the same as its original code for CULane.

\subsection{UFLD-v2.}
UFLD-v2~\cite{qin2022ultra}, an efficient row-wise method proposed in year 2022, is an improved version of UFLD~\cite{qin2020ultra}, which applies hybrid anchors of row and column to improve the predictions on side lanes. The classification dimensions are set to be 100 on column anchors, 200 on row anchors, and the number of anchors are 80 columns and 72 rows.

\subsection{CondLaneNet.}
CondLaneNet~\cite{liu2021condlanenet} is an effective and popular row-wise method. We apply the CondLaneNet-RIM version designed for forks lanes. Other settings remain the same as the provided code.

\subsection{BezierLaneNet.}
BezierLaneNet~\cite{feng2022rethinking} is a recent parameter-based method, which has a sensitive exploration capability on challenging lane cases. This method requires reconstructing the label to be parametric curves, e.g. bezier or polynomial curves. As SDLane has a high proportion of curves and a variaty of lane structures, we use 3 degree Bezier curves to reconstruct the label before training.

\subsection{CLRNet.}
CLRNet~\cite{zheng2022clrnet} is a state-of-the-art approach on TuSimple and CULane datasets using anchor-based model. We directly use the setting of the CULane's configuration file for SDLane, which performed better than TuSimple's configuration.



\end{document}